\documentclass{article}
\usepackage{iclr2027_conference,times}
\usepackage[T1]{fontenc}
\usepackage{CJKutf8}
\usepackage{amsmath,amssymb,amsthm}
\usepackage{graphicx}
\usepackage{xcolor}
\usepackage{capt-of}
\usepackage{booktabs}
\usepackage{float}
\usepackage{placeins}
\usepackage{tabularx}
\usepackage{longtable}
\usepackage{hyperref}
\usepackage{url}
\newtheorem{theorem}{Theorem}[section]
\newtheorem{proposition}[theorem]{Proposition}

\theoremstyle{definition}
\newtheorem{definition}[theorem]{Definition}
\theoremstyle{plain}
\providecommand{\Description}[1]{}
\iclrfinalcopy
\title{J-Miner: Recovering the Decision Logic of Fine-Tuned LLM Classifiers as Compact Rules}
\author{%
\parbox{\textwidth}{\raggedright
{\bf Yunfan Gao$^{1}$, Xinyi Huang$^{2}$, Tao Sheng$^{3}$, Haorui Song$^{1}$, Yun Xiong$^{2}$, Haofen Wang$^{4,*}$}\\[3pt]
{\normalfont\scriptsize \textsuperscript{1}Shanghai Research Institute for Intelligent Autonomous Systems, Tongji University \quad \textsuperscript{3}Meituan}\\
{\normalfont\scriptsize \textsuperscript{2}Shanghai Key Laboratory of Data Science, College of Computer Science and Artificial Intelligence, Fudan University}\\
{\normalfont\scriptsize \textsuperscript{4}College of Design and Innovation, Tongji University \quad \textsuperscript{*}Corresponding author}\\[1pt]
\resizebox{\textwidth}{!}{\normalfont\scriptsize\texttt{\{gaoyunfan1602,hwwysong,carter.whfcarter\}@gmail.com}, \texttt{\{xinyihuang25@m.,yunx@\}fudan.edu.cn}, \texttt{shengtao03@meituan.com}}%
}%
}
\hypersetup{
  hidelinks,
  pdftitle={J-Miner: Recovering the Decision Logic of Fine-Tuned LLM Classifiers as Compact Rules},
  pdfauthor={Yunfan Gao, Xinyi Huang, Tao Sheng, Haorui Song, Yun Xiong, Haofen Wang},
  pdfsubject={Executable Decision Compression and Mechanistic Interpretability of Language-Model Classifiers},
  pdfkeywords={language-model classifiers, executable decision compression, mechanistic interpretability, rule extraction}
}

\begin{document}

\maketitle
\begin{abstract}
Task-fine-tuned large language model (LLM) classifiers acquire task-specific decision knowledge, but this knowledge remains implicit in distributed internal computations, making their decision logic difficult to interpret.
We introduce the Executable Decision Compression (EDC) framework and propose J-Miner, which mines vocabulary-named variables from internal readouts and learns rules shared across inputs to produce executable explanations.
Analysis reveals that a small set of these variables captures much of the classifier's decision behavior, holding for both varying parameter scales within a family and distinct families.
Across six binary tasks, a rule using just one variable reproduces 76.7\% of source-classifier decisions on average, rising to 88.8\% with 16 variables.
Most of the decision information retained by these variables comes from internal activations beyond literal surface matching.
A lightweight text reader predicts the variable states, allowing the same fixed rules to execute independently of the source classifier.
\end{abstract}

\section{Introduction}

Fine-tuned large language model (LLM) classifiers inherit the language
understanding learned during pretraining to make task-specific judgments. Even
lightweight LLM classifiers can capture complex cues that fixed vocabularies and
handcrafted features miss, making them widely used in spam filtering, toxicity
detection, content moderation, and sentiment or intent analysis
\citep{caselli2021hatebert,lebail2026classifsae,labadietamayo2025distilling}.
Yet these classifiers remain black boxes. Each prediction outputs only a class
label, leaving the task-specific decision knowledge learned during fine-tuning
implicit inside the model and obscuring the decision logic that drives its
judgments.

Understanding this decision logic requires identifying both the internal
evidence behind a classifier's judgments and how that evidence is combined.
Probing methods test whether specific information can be decoded from internal
representations \citep{hewitt2019designing,gurnee2024language}, while sparse autoencoders (SAEs)
decompose these representations into sparse latent features
\citep{cunningham2023sparse}. However, probes evaluate predefined targets in
isolation, while feature decompositions disperse activations across thousands of
latent dimensions without specifying how they combine into task-level judgments.
We therefore propose the \emph{Executable
Decision Compression} (EDC) framework, built on the hypothesis that much of a
classifier's decision behavior may admit a compact representation over internal
evidence variables.
We examine this hypothesis through three research questions (RQs).
\textbf{RQ1 (Recover):} How compactly and faithfully can a classifier's
decisions be explained? \textbf{RQ2 (Reveal):} What internal evidence do these
explanations reveal, and how is it combined? \textbf{RQ3 (Execute):} Can
these explanations be executed independently to reproduce the classifier's
decisions?

\begin{figure}[!t]
  \centering
  \includegraphics[width=\textwidth]{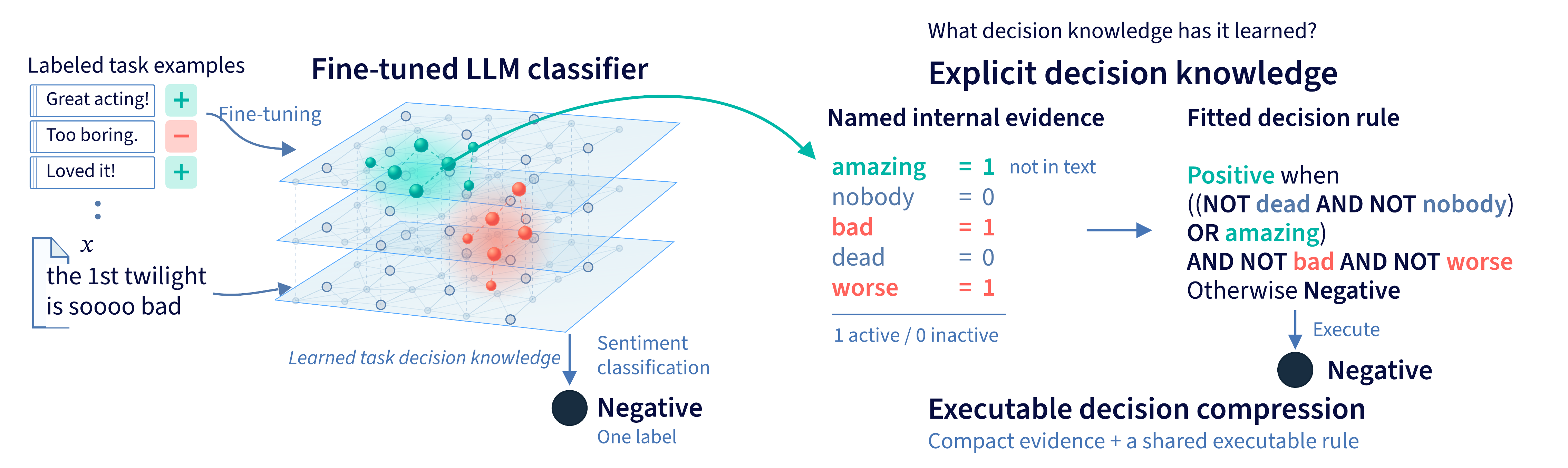}
  \caption{A fine-tuned LLM classifier outputs only a label, leaving
  decision logic implicit. J-Miner fits a shared executable rule over
  vocabulary-named internal readouts to reproduce its predictions.}
  \Description{A landscape diagram contrasts a fine-tuned classifier's output label on the left with explicit decision knowledge on the right. On the left, labeled sentiment examples feed a fine-tuning arrow into a layered LLM classifier. The input, the 1st twilight is soooo bad, receives the label Negative. A question above the right-hand representation asks what decision knowledge the classifier has learned. A connector from the model interior leads to a compact evidence state beside the fitted sentiment rule. The five vocabulary-named internal coordinates are amazing equals 1, nobody equals 0, bad equals 1, dead equals 0, and worse equals 1, where 1 means active and 0 means inactive. These are internal activations, not literal word detections. A not in text label beside amazing equals 1 highlights that its name is absent from the input. The fitted Boolean rule is ((NOT dead AND NOT nobody) OR amazing) AND NOT bad AND NOT worse. It predicts Positive when this condition is true and Negative otherwise. Executing the rule on the displayed state yields Negative because bad and worse are active, making the outer conjunction false. The footer below the right-hand representation names executable decision compression as compact evidence plus a shared executable rule. The layered model and training examples are schematic; the sentiment input, evidence state, and fitted rule match the Boolean replay in Figure 4b.}
  \label{fig:motivation}
\end{figure}

To instantiate EDC without training auxiliary feature dictionaries, J-Miner
projects the classifier's intermediate hidden states into its own vocabulary
space \citep{gurnee2026workspace} to obtain readable candidate events, while
the same variable selection and rule learning also apply to alternative
internal readouts. From these candidate events, J-Miner identifies a small
shared set of internal evidence variables and learns compact decision rules
from the frozen classifier's predictions to capture how these
variables combine into a judgment. Figure~\ref{fig:motivation} illustrates this
process using a fitted sentiment rule. Whereas the classifier
outputs only a \emph{Negative}
label, J-Miner evaluates a shared set of vocabulary-named variables such as
\emph{amazing}, \emph{nobody}, \emph{bad}, \emph{dead}, and \emph{worse}. Their states reflect
internal model activations rather than literal word matching alone. For
instance, both \emph{worse} and \emph{amazing} activate even though neither
word appears in the input text. A shared rule then combines these activation
states to reproduce the classifier's \emph{Negative} judgment. Expressing the
classifier's behavior through explicit variables and rules allows us to inspect
how internal evidence is combined behind a decision. Moreover, once the
variable definitions and rule are fixed, a lightweight text reader can predict
the required variable states to execute the same rule independently of the
source classifier.

We evaluate J-Miner across six binary and three multiclass tasks and
multiple model scales and families. For compact recovery (RQ1), measuring
behavioral fidelity across layer depths and variable budgets shows that a
tiny state budget captures most of a classifier's decision behavior, and that
this recoverability recurs across tasks, model families, and internal readout
protocols. Across the six binary tasks, a single internal variable already
captures 76.7\% of the source classifier's decisions on average, and 16
variables reach 88.8\% behavioral fidelity. Inspecting the revealed
explanations (RQ2) shows that non-literal internal events retain most of this
predictive signal even when a variable's name is absent from the input, and
rule-guided deletions shift the source classifier's margin in the predicted
direction more often than matched random deletions. Finally, for independent
execution (RQ3), holding the recovered rule fixed while supplying states from a
33M-parameter text reader separates reader-induced state changes from decision
changes. Standalone execution preserves 88.2\% and 87.9\% of source-state rule
decisions under state-only and joint supervision, respectively, while reaching
91.1\% and 93.3\% fidelity to the source classifier. The higher behavioral
fidelity under joint supervision accompanies more verdict-aligned score shifts,
rather than more faithful reconstruction of the source evidence states.
This work makes three contributions:
\begin{itemize}
  \item We propose the Executable Decision Compression framework and
  instantiate it with J-Miner to recover compact, shared decision rules over
  internal evidence events and inspect how those events combine into judgments.
  \item We formulate the minimum state-budget objective across observation
  depths and budgets, showing that compact recoverability recurs across tasks,
  model families, and internal observation protocols, with predictive signal
  carried primarily by non-literal internal activations.
  \item We execute the frozen rules with a standalone text reader and derive an
  exact margin and decision-change decomposition that separates reader-induced
  state discrepancies from behavioral changes relative to the source classifier.
\end{itemize}

\section{Executable Decision Compression}
\label{sec:edc}

\begin{figure*}[!t]
  \centering
  \includegraphics[width=\textwidth]{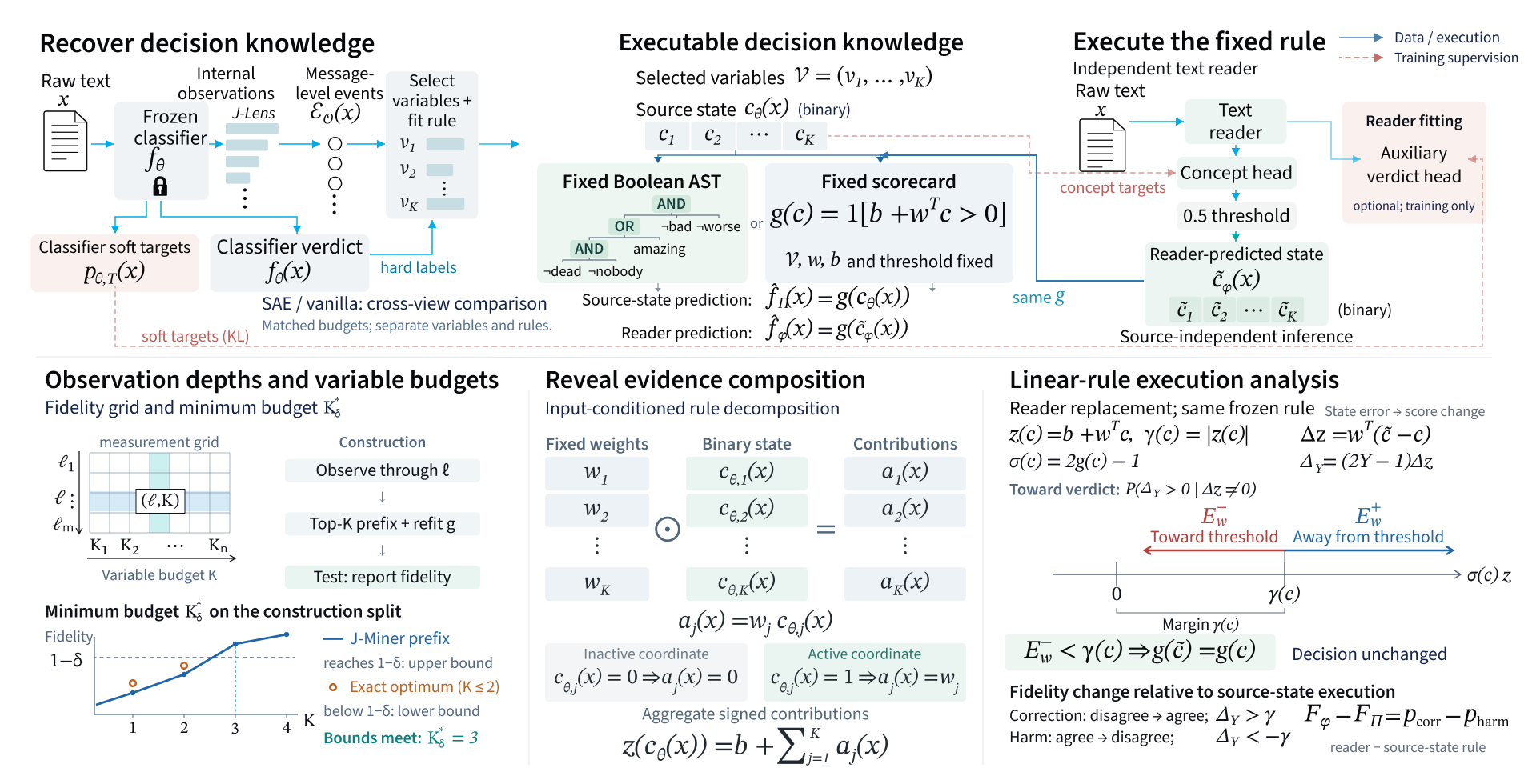}
  \caption{Overview of Executable Decision Compression (EDC) and J-Miner.}
  \Description{A method diagram organized around one executable decision
  knowledge object. Recover selects internal evidence variables and constructs
  a Boolean AST or linear scorecard using source-classifier predictions. Separate
  classifier verdict and soft-target boxes distinguish hard-label supervision for rule
  construction from soft targets for the auxiliary head.
  Source states and reader-predicted
  binary states feed the same fixed rule. The reader thresholds concept
  probabilities at 0.5; the optional auxiliary verdict output does not determine
  the final prediction. The depth and budget panel shows a
  measurement grid over observation depths and variable budgets with per-cell construction and evaluation.
  A construction-set schematic compares J-Miner prefix fidelity with exact
  optima for budgets up to two, yielding upper and lower bounds on the minimum
  budget $K^\star_\delta$; test data are used to report fidelity.
  Reveal multiplies fixed weights by input-conditioned binary states to obtain
  signed variable contributions, which sum with the intercept to form the
  linear rule score. Offline execution analysis relates
  reader-state errors to score changes and the rule margin. Toward verdict denotes
  the share of nonzero score changes toward the source classifier's verdict. Separate
  arrows show weighted errors toward and away from the threshold.
  $E_w^-<\gamma$ guarantees unchanged rule prediction. Corrections require
  $\Delta_Y>\gamma$ for a disagreeing source-state rule; harms require
  $\Delta_Y<-\gamma$ for an agreeing one. Their difference gives the net
  fidelity change relative to source-state execution.}
  \label{fig:pipeline}
\end{figure*}

We now formalize EDC and describe how J-Miner constructs and evaluates
executable decision rules from a frozen classifier, as illustrated in
Figure~\ref{fig:pipeline}.

\subsection{Problem Formulation}
\label{sec:edc-objective}

Consider a frozen source classifier $f_\theta:\mathcal X\to\mathcal Y$
that maps an input $x\sim\mathcal D$ to a prediction
via internal states $T_\theta(x)$. We present the binary formulation
$\mathcal Y=\{0,1\}$ below and extend it to multiclass settings in
Section~\ref{sec:edc-mining}.

Given an observation protocol $\mathcal O$ specifying how candidate internal
evidence is extracted, an executable decision representation
$\Pi=(\mathcal V,q_{\mathcal O,\mathcal V},g)$ selects $K$ ordered variables
$\mathcal V=(v_1,\ldots,v_K)$, reads their binary states from $T_\theta(x)$ via
$q_{\mathcal O,\mathcal V}$, and predicts the verdict via a shared rule $g$,
\begin{equation}
c_\theta(x)=q_{\mathcal O,\mathcal V}\big(T_\theta(x)\big)\in\{0,1\}^K,\qquad
\widehat f_\Pi(x)=g\big(c_\theta(x)\big).
\label{eq:edc-representation}
\end{equation}
While the state $c_\theta(x)$ varies by input, $\mathcal V$ and $g$ remain fixed
across $\mathcal D$. We define decision distortion as
$\varepsilon_\theta(\Pi)=\Pr_{X\sim\mathcal D}[\widehat f_\Pi(X)\neq f_\theta(X)]$
and behavioral fidelity to the source classifier as $1-\varepsilon_\theta(\Pi)$,
which serves as our primary evaluation target alongside gold-label task
accuracy. At low distortion, $\Pi$ forms a checkpoint-specific executable
approximation of $f_\theta$ on $\mathcal D$ whose rule specifies how the
extracted evidence combines into each prediction.

To keep this explanation compact and inspectable, EDC further seeks the smallest
variable budget $K=|\mathcal V|$ that preserves the source classifier's behavior
within a tolerated distortion $\delta$. Let $\mathcal A(\mathcal O,\mathcal G)$
denote the admissible representations with candidate variables from $\mathcal O$
and rules $g\in\mathcal G$. The minimum state budget is
\begin{equation}
K_\delta^\star(\mathcal O,\mathcal G)
=\min_{\Pi\in\mathcal A(\mathcal O,\mathcal G)}
\big\{|\mathcal V|:\varepsilon_\theta(\Pi)\le\delta\big\}.
\label{eq:edc-objective}
\end{equation}
Because computing $K_\delta^\star$ is NP-hard via a reduction from set cover
(Appendix~\ref{app:edc-small-budget}), J-Miner constructs $\mathcal V$ and $g$
heuristically on a construction split $D_c$ and evaluates them on a held-out
test set. For the empirical objective on $D_c$, any $K$-variable J-Miner rule
reaching fidelity $1-\delta$ yields an upper bound $K_\delta^\star\le K$.
Conversely, exhaustive search at $K\in\{1,2\}$ certifies $K_\delta^\star>K$
when optimal fidelity falls below $1-\delta$, and when the two bounds
meet, they determine the empirical $K_\delta^\star$ exactly.

\subsection{Mining Executable Decision Knowledge}
\label{sec:edc-mining}

To instantiate $\Pi=(\mathcal V,q_{\mathcal O,\mathcal V},g)$, J-Miner maps
intermediate states $T_\theta(x)$ into discrete candidate events through an
observation protocol $\mathcal O$. As our default instantiation, we project
hidden states into the classifier's vocabulary space via the Jacobian lens
(J-Lens) \citep{gurnee2026workspace}, obtaining token-named event coordinates
without training auxiliary feature dictionaries so that internal activations can
be compared directly against input words. For a residual
state $h_{\ell,p}(x)\in T_\theta(x)$ at layer $\ell$ and token position $p$,
J-Lens applies an average Jacobian $J_\ell$ calibrated on external text to
transport the state into the final residual space, followed by the classifier's
final normalization and unembedding $U$ to produce vocabulary scores
$r_{\ell,p}(x)=U\,\mathrm{Norm}(J_\ell h_{\ell,p}(x))$. To test whether
small-budget recoverability is a property of the classifier's internal states
rather than a specific readout operator, we apply the identical variable
selection and rule fitting to vanilla logit-lens readout and task-trained sparse
autoencoder (SAE) latents.

To construct message-level variables, we retain the top $R$ vocabulary
coordinates of $r_{\ell,p}(x)$ in the designated layers at content positions,
defined as input tokens excluding the prompt template and decision position,
decode and normalize their names, and exclude shared template terms together with task-
and class-label words. We take the union of the retained names across
layers and positions to form the event set $\mathcal E_{\mathcal O}(x)$.
Readouts with the same normalized name merge into a single message-level event
regardless of the layer or position at which they occur. On the construction
split $D_c$, we discard candidates below a minimum support threshold and rank
the remainder by their class-conditional trigger-rate gap under the frozen
classifier's predictions, $\Delta(v)=|\pi_1(v)-\pi_0(v)|$ where
$\pi_y(v)=\Pr_{x\in D_c}[v\in\mathcal E_{\mathcal O}(x)\mid f_\theta(x)=y]$,
extending this score to multiclass tasks via the one-vs-rest gap
$\max_y|\pi_y(v)-\pi_{\neg y}(v)|$. Selecting the top $K$ names yields the ordered variable set
$\mathcal V=(v_1,\ldots,v_K)$ and defines the binary state map
$q_{\mathcal O,\mathcal V}$ through
\begin{equation}
c_{\theta,j}(x)=\big[q_{\mathcal O,\mathcal V}(T_\theta(x))\big]_j
=\mathbf 1\big[v_j\in\mathcal E_{\mathcal O}(x)\big],\qquad j\in\{1,\ldots,K\},
\label{eq:edc-state}
\end{equation}
where $c_{\theta,j}(x)=1$ records that $v_j\in\mathcal E_{\mathcal O}(x)$ and
$c_{\theta,j}(x)=0$ records that $v_j$ is not observed under $\mathcal O$.
Here, vocabulary names index protocol-defined readout events rather than
ordinary lexical meanings; an absent state $c_{\theta,j}(x)=0$, written as
$\textsc{not}\ v_j$ in Boolean rules, denotes non-observation under
$\mathcal O$ rather than the semantic negation of the printed word. Over these binary
states $c_\theta(x)$, we fit a shared rule $g\in\mathcal G$ to imitate
$f_\theta(x)$ on $D_c$ using two complementary rule families. A bounded-size
Boolean abstract syntax tree (AST) combines active and
absent variable states through \textsc{and} and \textsc{or} to
capture discrete logical interactions, selected by construction-split fidelity.
A linear scorecard captures additive evidence accumulation with signed weights
$w\in\mathbb R^K$ and intercept $b\in\mathbb R$:
\begin{equation}
g(c)=\mathbf 1\big[z(c)>0\big],\qquad
z(c)=b+w^\top c,\qquad
a_j(x)=w_j c_{\theta,j}(x).
\label{eq:edc-scorecard}
\end{equation}
We fit $(w,b)$ via class-balanced L2-regularized logistic regression against
$f_\theta(x)$, or via multinomial logistic regression with $\arg\max_y z_y(c)$
for multiclass tasks. After fitting on $D_c$, we freeze
$\mathcal V$, $q_{\mathcal O,\mathcal V}$, and $g$. For any held-out input $x$,
the frozen tuple $\Pi$ produces an explicit execution trace. The AST identifies
the exact conjunction or disjunction of active and absent variables that triggers
the rule's verdict, while the linear rule decomposes $z(c_\theta(x))$ into per-variable
signed contributions $a_j(x)$ that predict how toggling or replacing a
variable shifts the rule's decision score.

\subsection{Observation Depths and Variable Budgets}
\label{sec:edc-measurement}

We measure how behavioral fidelity varies jointly with internal
observation depth and variable budget across a grid of cutoff layers and
budgets. A measurement protocol $\mathcal P$ fixes the source classifier,
position and vocabulary filtering, candidate ranking, and rule fitting
procedure, while varying the cumulative layer window. For a cutoff layer $\ell$,
the observation protocol $\mathcal O_{\ell_0:\ell}$ pools candidate events
across content positions from layer $\ell_0$ through $\ell$. At each cutoff
$\ell$, J-Miner ranks the support-filtered candidates once on the construction
split $D_c$. Each budget $K$ takes the top-$K$ prefix $\mathcal V_{\ell,K}$,
forms the binary state $c_\theta^{(\ell,K)}(x)$ under $\mathcal O_{\ell_0:\ell}$,
and independently fits a rule $g_{\ell,K}$. Evaluating every pair
$(\ell,K)$ on the same held-out test set $\mathcal T$ defines the fidelity grid:
\begin{equation}
\widehat F_{\mathcal P}(\ell,K)
=\frac{1}{|\mathcal T|}\sum_{x\in\mathcal T}
\mathbf 1\Big[g_{\ell,K}\big(c_\theta^{(\ell,K)}(x)\big)=f_\theta(x)\Big].
\label{eq:dcs}
\end{equation}

At fixed cutoff $\ell$, varying $K$ traces the tradeoff between explanation
size and behavioral fidelity. At fixed budget $K$, varying $\ell$ tests
whether compact rules can use accumulated evidence at content positions to
recover the classifier's predictions before label readout at the decision
position stabilizes.

\subsection{Standalone Rule Execution}
\label{sec:edc-transfer}

To test whether the recovered rule can operate as a self-contained decision
mechanism without access to the source classifier's internal states, we fix the
variable definitions, their order, and the rule parameters, and train an
independent text reader $s_\phi:\mathcal X\to[0,1]^K$ to predict $c_\theta(x)$
directly from the input text. The reader is supervised by source states
$c_\theta(x)$ (J-only), while the frozen rule $g$ remains outside the training
graph. In a second variant (J+Aux), the encoder also feeds a training-only
auxiliary head that matches the source classifier's soft verdict distribution
$p_{\theta,T}(x)$ via temperature-scaled KL divergence. At inference, we discard any
auxiliary head, binarize $s_\phi(x)$ coordinate-wise at $0.5$, and execute the
frozen rule:
\begin{equation}
\widetilde c_\phi(x)=\mathbf 1\big[s_\phi(x)\ge0.5\big],\qquad
\widehat f_\phi(x)=g\big(\widetilde c_\phi(x)\big).
\label{eq:edc-student}
\end{equation}

Whether state errors alter a binary rule decision depends on their signed
weights relative to the rule margin $\gamma(c)=|z(c)|$ and decision sign
$\sigma(c)=2g(c)-1\in\{-1,+1\}$. Replacing $c$ with $\widetilde c$ changes
coordinate $j$'s contribution by $\delta_j=w_j(\widetilde c_j-c_j)$. With
$(u)_+=\max(u,0)$, this separates into the adverse perturbation
$E_w^-(c,\widetilde c)=\sum_j(-\sigma(c)\,\delta_j)_+$ toward the threshold and
the favorable perturbation $E_w^+(c,\widetilde c)=\sum_j(\sigma(c)\,\delta_j)_+$
away from it. Because the
replaced score satisfies
\begin{equation}
\sigma(c)\,z(\widetilde c)
=\gamma(c)+E_w^+(c,\widetilde c)-E_w^-(c,\widetilde c),
\label{eq:edc-guarantee}
\end{equation}
any state replacement with $E_w^-(c,\widetilde c)<\gamma(c)$ preserves the rule
decision, $g(\widetilde c)=g(c)$.

Across $X\sim\mathcal D$ with source verdict $Y=f_\theta(X)$, source state
$c=c_\theta(X)$, reader state $\widetilde c=\widetilde c_\phi(X)$, and
verdict-aligned score shift $\Delta_Y=(2Y-1)(z(\widetilde c)-z(c))$, the change
from source-state fidelity $F_\Pi=1-\varepsilon_\theta(\Pi)$ to standalone
fidelity $F_\phi=\Pr[\widehat f_\phi(X)=Y]$ decomposes into \emph{corrections}
$p_{\mathrm{corr}}$ minus \emph{harms} $p_{\mathrm{harm}}$ whenever
$z(\widetilde c)\neq 0$:
\begin{equation}
F_\phi-F_\Pi
=\Pr\big[g(c)\neq Y,\ \Delta_Y>\gamma(c)\big]
-\Pr\big[g(c)=Y,\ \Delta_Y<-\gamma(c)\big].
\label{eq:edc-decomposition}
\end{equation}
Replacement corrects a disagreeing rule decision when
$\Delta_Y>\gamma(c)$ and harms an agreeing one when
$\Delta_Y<-\gamma(c)$
(Appendices~\ref{app:edc-guarantee-formal}--\ref{app:edc-corrections}).

\section{Experimental Setup}
\label{sec:setup}

We evaluate J-Miner on six binary text-classification tasks, namely SMS,
Sentiment, Formality, IMDB, Toxicity, and Sarcasm, and on three multiclass
tasks, HateXplain, SNIPS-3, and SNIPS-7. Each task uses an independently
fine-tuned, frozen Qwen3.5-0.8B classifier. A four-task panel adds Qwen3.5-2B
and -4B, Gemma-3-1B, Llama-3.2-1B, and Phi-4-mini. For each classifier, J-Lens
is calibrated on external WikiText-103 text. Source-classifier predictions on
the construction split $D_c$ supervise variable selection and rule fitting. All selected variables
and rules are frozen before evaluation on the disjoint held-out test set. To compare decision
evidence from internal concepts and literal words, J-Miner rules and the
Surface baseline share data splits, source-classifier targets, variable budgets
$K$, and the linear rule-fitting procedure, differing only in where their
indicators come from. The primary metric is source-classifier fidelity, the
agreement between final predictions and the frozen source classifier's
predictions, while gold accuracy measures agreement with dataset labels and
separately reflects task performance. See Appendices~\ref{app:rq1-protocol}--\ref{app:reader-replacement} for full details.

For standalone rule execution on the six binary tasks, we keep the recovered
variables and rule fixed and instantiate the text reader $s_\phi$ of
Section~\ref{sec:edc-transfer} with a 33.2M-parameter E5-small-v2 encoder. We
train the J-only and J+Aux readers under state-only and joint soft-verdict
supervision to examine how reader-induced state changes interact with the
frozen rule. To benchmark encoder capacity, an unconstrained
Plain KD baseline fits a direct classification head on the same E5-small-v2
encoder, data splits, and optimization settings without concept states or rules. J-only uses state
reconstruction as its sole training loss; checkpoint selection for all reader
variants uses inner-development agreement with the source classifier. Because
the rule is fixed, we evaluate standalone execution jointly by concept macro-F1
against source states, the share of test inputs certified by the margin
condition $E_w^-<\gamma$ of Equation~\eqref{eq:edc-guarantee}, agreement with
source-state rule decisions, and the decomposition of behavioral changes into
corrections and harms alongside source-classifier fidelity.

\section{Results}
\par
\begin{figure}[!t]
  \centering
  \includegraphics[width=\textwidth,trim=0 12bp 0 0,clip]{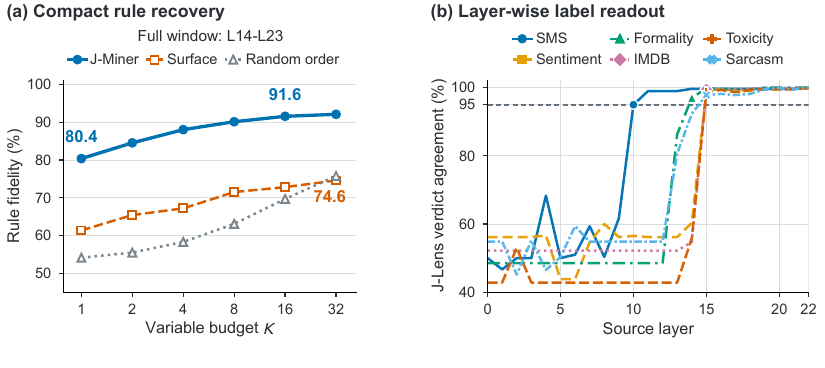}
  \caption{Compact rule recovery and layer-wise label readout.
  (a) Held-out fidelity to the source classifier, averaged over five tasks.
  (b) Held-out J-Lens verdict agreement with final predictions.}
  \Description{The left panel plots five-task mean source-classifier fidelity
  against variable budgets from 1 to 32 for internal variables in J-Miner's
  ranking, random orderings of the same candidate pool, and Surface words. One internal variable
  reaches 80.4 percent, exceeding 74.6 percent for 32 Surface words; 16 internal
  variables reach 91.6 percent. The right panel shows six held-out layer-wise J-Lens
  label-agreement curves from L0 to L22. Sustained 95-percent agreement begins
  at L10 for SMS, L14 for Formality, and L15 for Sentiment, IMDB,
  Toxicity, and Sarcasm.}
  \label{fig:compression-evidence}
\end{figure}

\noindent\phantomsection\label{sec:rq1-recover}%
\textbf{RQ1: Compact recovery of classifier decisions.}
We first test whether a fine-tuned classifier's decisions can be compressed into
compact rules over shared internal variables across budgets $K$. Across the six binary tasks in
Table~\ref{tab:recovery}, fine-tuning raises gold accuracy from
50.0\%--76.2\% for Zero-shot to 85.5\%--98.0\% for the fine-tuned Source
classifier, whereas the Majority baseline matches only 50.0\%--57.2\%
of the classifier's predictions. Against these baselines, a single internal variable with an
intercept already captures 76.7\% of the source classifier's decisions on
average, and 16 variables raise mean behavioral fidelity to 88.8\%.
Figure~\ref{fig:compression-evidence}a traces this budget curve from $K=1$ to
$32$ against literal-word Surface rules across the five binary tasks with
at least 32 support-filtered surface words, excluding Sarcasm with 24. On these five
tasks, one J-Miner variable reaches 80.4\% fidelity, surpassing 1-word and 32-word Surface rules at 61.4\%
and 74.6\%, while 16 variables reach 91.6\% and exceed
Surface across $K=1$ to $32$ on all five tasks by 19.0~pp on average.

Ranking by $\Delta(v)$ orders candidates by their best achievable
single-variable class-balanced fidelity on the construction split. At $K=2$, J-Miner reaches 81.1\% mean held-out fidelity across
the six binary tasks, within 1.0~pp of exhaustively selected
internal pairs at 82.1\% and well above exhaustively selected
literal-word pairs at 65.8\%.
For the J-Lens protocol and linear rule family, coinciding bounds certify
$K_{0.10}^\star=1,2,3$ for SMS, Toxicity, and Formality, respectively, at
90\% construction-split fidelity. At $K=16$, the ranked rules exceed the median
held-out fidelity of 100 random orderings of the support-filtered internal pool
by 20.6~pp.

Across all nine binary and multiclass tasks in Table~\ref{tab:recovery}, both
Boolean AST rules with at most five literals
and 16-variable LR scorecards exceed the Majority baseline and Surface rules in
behavioral fidelity, with LR exceeding
Surface by 6.0--29.5~points on every binary task. On 281 held-out binary
examples where the source classifier's prediction differs from the gold
label, J-Miner follows 66.2\% of its decisions compared with 50.9\% for
Surface. Refitting the rule weights to gold labels while retaining the
same variables selected using source predictions yields 88.7\% mean
fidelity and follows 63.7\% of these errors.

\begin{table}[htbp]
  \caption{Recovering classifier decisions with compact rules.}
  \label{tab:recovery}
  \label{tab:rq1-recovery}
  \medskip
  \centering
  \small
  \renewcommand{\arraystretch}{1.10}
  \setlength{\tabcolsep}{1.6pt}
  \begin{tabular*}{\textwidth}{@{\extracolsep{\fill}}llrrrrrr@{\hspace{8pt}}rrr@{}}
    \toprule
    & & \multicolumn{6}{c}{Binary} & \multicolumn{3}{c}{Multiclass} \\
    \cmidrule(lr){3-8}\cmidrule(lr){9-11}
    Method & \shortstack{Metric\\(\%)} & SMS & Sentiment & Formality & IMDB & Toxicity & Sarcasm & HateXplain & SNIPS-3 & SNIPS-7 \\
    \midrule
    Source & Acc. & 98.0 & 90.5 & 94.8 & 94.5 & 85.5 & 88.8 & 62.2 & 100.0 & 98.2 \\
    Zero-shot & Acc. & 51.3 & 66.8 & 54.3 & 76.2 & 60.7 & 50.0 & 35.2 & 69.4 & 74.6 \\
    Majority & Fid. & 50.0 & 56.2 & 51.5 & 52.2 & 57.2 & 54.8 & 47.7 & 33.3 & 14.0 \\
    \midrule
    Surface & Fid. & 92.3 & 62.3 & 78.7 & 69.8 & 61.0 & 66.3 & 65.3 & 98.0 & 81.7 \\
    \addlinespace[2pt]
    AST & Acc. & 96.7 & 80.7 & 91.2 & 82.5 & \textbf{84.3} & 68.8 & 51.2 & 98.7 & \textbf{85.1} \\
          & Fid. & 98.0 & 81.8 & 94.0 & 83.3 & 89.5 & 69.3 & \textbf{67.3} & 98.7 & \textbf{86.5} \\
    \addlinespace[2pt]
    LR & Acc. & \textbf{98.3} & \textbf{80.8} & \textbf{92.5} & \textbf{88.0} & \textbf{84.3} & \textbf{73.8} & \textbf{52.3} & \textbf{99.3} & 85.0 \\
          & Fid. & \textbf{98.3} & \textbf{84.3} & \textbf{95.3} & \textbf{89.5} & \textbf{90.5} & \textbf{75.0} & 66.0 & \textbf{99.3} & 86.3 \\
    \bottomrule
  \end{tabular*}

  \medskip
  \begin{minipage}{\textwidth}
    \footnotesize
    Acc.: gold accuracy; Fid.: source-classifier fidelity. AST: at most five
    literals per binary rule or per class, selected from the same 16-variable
    candidate set; LR/Surface: 16 input variables.
  \end{minipage}
\end{table}

Varying the observation depth shows that content-position readouts support
compact rule recovery at L14. On Sentiment, IMDB, and Toxicity, 16-variable
rules using L14 alone reach 83.0\%--89.0\% held-out fidelity, within
1.8~points of the full L14--L23 window. By contrast, direct label readout at
the terminal prompt position in Figure~\ref{fig:compression-evidence}b agrees
with the source classifier on 54.7\%--60.3\% of held-out inputs at L14 and
reaches sustained 95\% agreement at L15 (Appendix~\ref{app:dcs-grid}).

\begin{figure}[htbp]
  \centering
  \includegraphics[width=\textwidth]{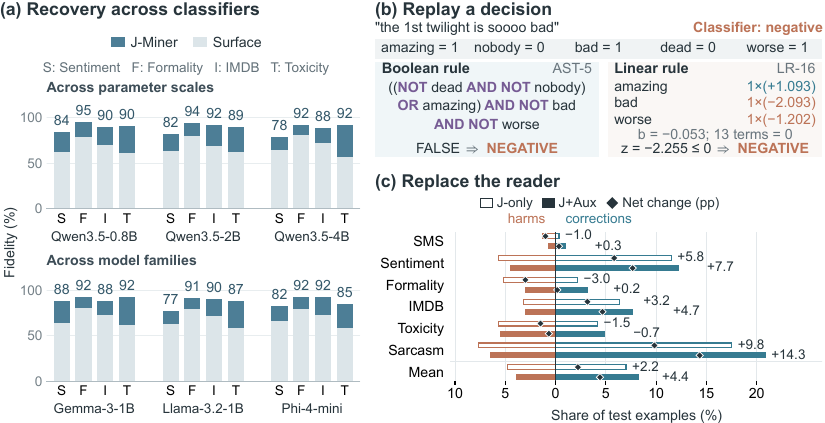}
  \caption{Recovering and executing decision rules.
  (a) Source fidelity (pale: Surface).
  (b) Boolean and linear replay.
  (c) Reader replacement (diamonds: net change).
  Mean weights tasks equally.}
  \Description{The first panel separates three Qwen3.5 model sizes from Gemma,
  Llama, and Phi classifiers in two vertically arranged zero-baseline bar plots.
  Each model has four task bars: Sentiment, Formality, IMDB, and Toxicity.
  Muted blue bars show J-Miner fidelity from zero, with pale gray-blue Surface
  bars overlaid from the same baseline. The exposed blue portion is their difference.
  The 24 J-Miner fidelity values are labeled as rounded whole percentages.
  The second shares five named indicator states across a Boolean AST-5 rule
  and an LR-16 calculation. Purple operators distinguish Boolean composition
  from concept names. The final frozen Boolean rule evaluates to false, and
  the linear score is minus 2.255; both reproduce the source classifier's negative verdict.
  The third retains
  correction and harm bars, with all 14 net-change labels beside the bar ends
  for two readers on six binary tasks and their mean. Recovery fits a separate rule for each task and
  classifier; reader replacement holds each task's rule fixed.}
  \label{fig:result-panels}
  \label{fig:rq1-generality}
  \label{fig:rq1-sentiment-case}
  \label{fig:correction-harm}
\end{figure}

This small-budget recoverability recurs across tasks, model scales and
families, and internal observation protocols $\mathcal O$. In
Figure~\ref{fig:result-panels}a, across four tasks spanning three Qwen sizes and
Gemma, Llama, and Phi classifiers, 16-variable rules reach 77.3\%--95.3\%
fidelity and exceed Surface in all 24 cells by 11.5--35.5~pp. Under the identical selection and
fitting procedure, mean fidelity is 89.92\% for J-Lens, 90.83\% for vanilla
readout, and 91.71\% for SAE latents versus 67.96\% for Surface, showing that
compact recoverability is not specific to Jacobian transport
(Appendix~\ref{app:rq2-interfaces}). Both J-Lens and vanilla readout avoid
dictionary training and expose vocabulary-indexed events, while $J_\ell$
calibration yields complete English words for 63 of 64 selected variables
versus 24 of 64 for vanilla, whose other 40 are 29 non-English script tokens
such as Arabic and CJK and 11 subword fragments.

\par\noindent\phantomsection\label{sec:rq2-reveal}%
\textbf{RQ2: Internal evidence and its composition.}
We next examine what internal evidence the recovered 16-variable rules capture
and how that evidence combines into decisions. The selected variables include
recognizable task cues, such as \texttt{click} and \texttt{cash} in SMS and
\texttt{happy} and \texttt{bad} in Sentiment. Across the six binary tasks, all
96 selected variables retain their construction-split direction of association
with the source classifier's predictions on held-out inputs. These variables activate primarily when
their printed names are absent from the input: literal matches account for only
9.8\% of activations on average across the six binary tasks. Non-literal activations
supply 73\%--99\% of the class trigger-rate gap, averaging 90.5\%, and refitting
rules on non-literal activations alone reaches 88.6\% mean held-out
fidelity versus 88.8\% with all activations, whereas literal activations alone
drop to 63.3\%, below Surface at 71.7\% (Appendix~\ref{app:rq2-nonliteral}).

A variable's printed name therefore identifies an internal readout event whose
activation can occur without a literal match. To examine its dependence on
readout positions, we exclude positions without letters or digits from pooling
while leaving the input and hidden states unchanged. Frozen-rule fidelity
decreases from 84.3\% to 80.5\% in Sentiment and from 90.5\% to 84.7\% in
Toxicity, remaining above Surface at 62.3\% and 61.0\%. The remaining content-position readouts
thus retain information used by the recovered rules.

Across tasks, the recovered rules combine these internal cues either by
accumulating aligned evidence, such as promotional cues in SMS, or by weighing
competing signals against one another.
On the Sentiment input ``the 1st twilight is soooo bad'' in
Figure~\ref{fig:result-panels}b, the linear scorecard combines the
positive-weight activation \texttt{amazing}, read non-literally at
\emph{soooo}, with stronger negative contributions from \texttt{bad} and
\texttt{worse}. The Boolean AST-5 rule instead rejects the positive verdict
because the required absence conditions for \texttt{bad} and \texttt{worse}
are violated. Both rules reproduce the source classifier's negative verdict.

To assess whether readout-localized words are behaviorally relevant to the
source classifier, we select the active variable with the largest absolute rule
contribution and fix the predicted edit direction to
$\operatorname{sign}(-w_j c_{\theta,j}(x))$ before editing. We delete the words
aligned to its readout positions and compare the source-classifier margin change
with random deletions of the same number of words from the same input. Among
inputs satisfying the edit-validity filters, rule-guided deletion increases
direction agreement by 6.73--21.46~pp across the six tasks,
averaging 12.45~pp with every paired 95\% confidence interval
above zero (Appendix~\ref{app:rq2-text-edits}). For
instance, on a Toxicity input where J-Lens reads \texttt{hate} non-literally at
\emph{dark} and \emph{shameless}, deleting those two words deactivates
\texttt{hate} along with three other active variables and flips both the source
classifier margin from $+7.00$ to $-6.13$ and the rescored rule from $+4.24$ to $-1.31$
toward non-toxic, whereas randomly deleting \emph{won} and \emph{best} leaves the
internal state and both toxic verdicts intact. Because a text edit can change
multiple evidence variables, this test evaluates rule-guided behavioral
relevance rather than the isolated causal effect of a single variable.

\par\noindent\phantomsection\label{sec:rq3-transfer}%
\textbf{RQ3: Standalone execution of recovered rules.}
Holding each task's 16 variables and fitted rule weights fixed while
supplying their states $\widetilde c_\phi(x)$ with a 33M-parameter text reader
$s_\phi$ allows standalone execution without the 0.8B source classifier and
separates reader-induced state changes from behavioral changes. Across the six
binary tasks in Table~\ref{tab:transfer}, optimizing $s_\phi$ with the
concept-state loss alone in J-only reaches 91.1\% mean behavioral fidelity,
compared with 88.8\% on source states $g(c_\theta)$, while joint soft-verdict
supervision in J+Aux raises mean fidelity to 93.3\%.
J+Aux computes every final verdict through the frozen rule, whereas
Plain KD uses a direct classification head on the same encoder and reaches
92.8\% mean fidelity. Paired comparisons detect no significant fidelity
difference on any task, as reported in Appendix~\ref{app:reader-replacement}.
At the same time, concept macro-F1 falls from .686 under J-only to .655 under
J+Aux, showing that higher behavioral fidelity does not imply better overall
reconstruction of the source evidence states.

\begin{table}[htbp]
  \caption{Standalone rule execution on text-predicted concept states.}
  \label{tab:transfer}
  \medskip
  \centering
  \small
  \renewcommand{\arraystretch}{1.10}
  \setlength{\tabcolsep}{3pt}
  \begin{tabular*}{\textwidth}{@{\extracolsep{\fill}}lrrrrrrrrrr@{}}
    \toprule
    & \multicolumn{4}{c}{Source-classifier fidelity (\%)}
    & \multicolumn{2}{c}{Concept F1}
    & \multicolumn{2}{c}{Cover (\%)}
    & \multicolumn{2}{c}{Keep (\%)} \\
    \cmidrule(lr){2-5}\cmidrule(lr){6-7}\cmidrule(lr){8-9}\cmidrule(lr){10-11}
    Task & $g(c_\theta)$ & J-only & J+Aux & Plain KD & J-only & J+Aux & J-only & J+Aux & J-only & J+Aux \\
    \midrule
    SMS       & 98.3 & 97.3 & 98.7 & 98.3 & .919 & .914 & 98.0 & 98.3 & 98.3 & 98.3 \\
    Sentiment & 84.3 & 90.2 & 92.0 & 91.3 & .571 & .564 & 79.8 & 81.0 & 82.8 & 83.3 \\
    Formality & 95.3 & 92.3 & 95.5 & 95.5 & .738 & .696 & 89.3 & 93.5 & 92.7 & 93.8 \\
    IMDB      & 89.5 & 92.7 & 94.2 & 93.2 & .738 & .721 & 86.2 & 88.2 & 90.5 & 89.3 \\
    Toxicity  & 90.5 & 89.0 & 89.8 & 88.5 & .682 & .672 & 88.7 & 88.3 & 90.2 & 89.7 \\
    Sarcasm   & 75.0 & 84.8 & 89.3 & 90.2 & .466 & .363 & 61.2 & 69.7 & 74.8 & 72.7 \\
    \midrule
    Mean      & 88.8 & 91.1 & 93.3 & 92.8 & .686 & .655 & 83.9 & 86.5 & 88.2 & 87.9 \\
    \bottomrule
  \end{tabular*}

  \medskip
  \begin{minipage}{\textwidth}
    \footnotesize
    Plain KD: unconstrained direct-head reference on the same encoder; Cover: paired share with $E_w^-(c_\theta,\widetilde c_\phi)<\gamma(c_\theta)$; Keep: agreement with $g(c_\theta)$.
  \end{minipage}
\end{table}

To understand how the frozen rules maintain high fidelity when executed on
text-predicted states, we first compare the reader states $\widetilde c_\phi(x)$
against the source states $c_\theta(x)$ and examine why state prediction errors
rarely flip the rule decision. Across test inputs in Table~\ref{tab:transfer},
J-only and J+Aux reach concept macro-F1 scores of .686 and
.655, matching all 16 source coordinates on 18.4\% and 17.1\% of examples
and differing on 3.7 and 4.2 variables on average. Despite these
coordinate-level discrepancies, the frozen rules retain 88.2\% and 87.9\%
of source-state decisions under Keep.
Equation~\eqref{eq:edc-guarantee} accounts for this stability because a rule
decision flips only when adverse state perturbations $E_w^-$ exceed the
source margin $\gamma(c)$, and the one-sided condition $E_w^-<\gamma$
reported as Cover certifies 83.9\% of test inputs for
J-only and 86.5\% for J+Aux against decision flips.

We next examine why standalone execution reaches higher source-classifier
fidelity than source-state execution even as
J+Aux shows lower concept macro-F1 than J-only. By
Equation~\eqref{eq:edc-decomposition}, a verdict-aligned score shift cannot harm
an agreeing rule decision and corrects a disagreeing decision only when it
crosses the decision boundary. Empirically, the text readers exhibit strong
alignment of this kind:
whenever reader replacement changes the rule score, 86.1\% of score shifts under
J-only and 91.0\% under J+Aux move toward the source
classifier's verdict. The additional
state differences under J+Aux reinforce the classifier's verdict,
raising corrections from 7.0\% to 8.3\% of test inputs while lowering harms from
4.8\% to 3.9\% in Figure~\ref{fig:correction-harm}c. Across individual tasks,
J+Aux significantly improves over source-state execution on the three
tasks with the lowest initial fidelity, namely Sentiment, IMDB, and Sarcasm,
with exact McNemar $p<.001$, while remaining within 0.7~pp on the other three.

The frozen 16-variable rule shows how reader-predicted states affect each
decision. On the Sentiment input in Figure~\ref{fig:rq1-sentiment-case}b,
J-only drops the non-literal positive cue \texttt{amazing} and activates seven
additional negative variables, shifting the score from $-2.255$ to $-8.782$
while retaining the negative verdict.
Thus, source-state execution $g(c_\theta(x))$ characterizes the decision logic
supported by the classifier's own activations, whereas standalone execution
$g(\widetilde c_\phi(x))$ evaluates rule stability under text-predicted states.
Holding the rule fixed separates verdict agreement driven by verdict-aligned
state shifts $\Delta_Y>0$ from faithful recovery of $c_\theta(x)$.

\section{Conclusion}

Through the Executable Decision Compression (EDC) framework and J-Miner, we
showed that the decisions of fine-tuned LLM classifiers can be recovered under
a small state budget over shared internal variables and an explicit rule. This
compact recoverability recurs across the evaluated tasks, model scales and
families, and internal observation protocols, while consistently outperforming
equally sized literal-word rules. The recovered rules reveal that predictive
signal is carried primarily by non-literal internal events and show how those
events combine into judgments. Finally, holding the recovered rule fixed during
standalone text-reader execution separates reader-induced state changes from
behavioral changes, turning implicit classifier behavior into explanations that
can be both executed and inspected.

\section{Related Work}

Existing interpretability methods examine the internal evidence behind a
classifier's predictions either by localizing influential signals across inputs
and hidden pathways or by mapping intermediate activations into vocabulary
distributions and sparse feature directions. Along the first line, input
attribution traces salient tokens \citep{sundararajan2017axiomatic,abnar2020quantifying},
probes and concept vectors test whether target properties are linearly decodable
from hidden states \citep{hewitt2019designing,kim2018tcav}, and causal circuit
analyses isolate the internal subgraphs that mediate model behavior
\citep{geiger2021causal,marks2025sparse}. To inspect hidden states in the
model's vocabulary, lens methods apply the unembedding layer directly
\citep{nostalgebraist2020logitlens}, after learned affine maps
\citep{belrose2023tunedlens}, or through an average Jacobian projection to the
final residual space \citep{gurnee2026workspace}. Sparse autoencoders (SAEs)
instead decompose activations into learned dictionary directions
\citep{cunningham2023sparse}, and ClassifSAE jointly trains an SAE with a linear
head to classify from sparse latents \citep{lebail2026classifsae}. Standard SAE-based classifiers require auxiliary dictionary training, activate
input-varying subsets from thousands of directions, and produce unlabeled latent
indices requiring post-hoc naming; within our framework, EDC imposes the same
small shared budget $K$ on both vocabulary readouts and learned SAE latents
under a common rule-fitting protocol.

Classical rule extraction approximates a trained black-box predictor by fitting
trees or symbolic rules to its input-output behavior
\citep{craven1995extracting,bastani2017interpreting,frosst2017distilling,ribeiro2018anchors},
distinguishing agreement with the source model from task accuracy
\citep{jagielski2020high}. To tie rules to model internals, \mbox{NEUROLOGIC}
synthesizes logic rules over thresholded neurons and grounds them through
post-hoc keyword templates \citep{geng2025neurologic}, while MechaRule localizes
sparse neuron circuits whose ablation disrupts rule behavior
\citep{sovrano2026mecharule}. Concept bottleneck models instead route
predictions through an explicit concept layer during training or distillation
\citep{koh2020concept,oikarinen2023labelfree,yang2023labo,ludan2023interpretable,%
rao2024discoverthenname,labadietamayo2025distilling,sun2025cbllm,hsieh2023distilling},
though learned concept representations can suffer from semantic misalignment and
label leakage \citep{margeloiu2021intended,mahinpei2021promises}. In contrast,
we recover compact rules post hoc over vocabulary-indexed variables under a
controlled budget, and hold both the variables and rule fixed during standalone
execution to decompose how reader-induced state changes preserve, correct, or
harm the recovered rule's agreement with the source classifier
(\hyperref[sec:rq3-transfer]{RQ3}).

\clearpage
\subsection*{AI use statement}

All core research in this work, including problem formulation, concept and
method design, theoretical formulation and proofs, experimental execution,
analysis of results, and scientific conclusions, was conceived and carried out
by the authors. Generative AI tools were used as auxiliary aids for language
polishing, proofreading, editorial feedback, checking notation and derivation
steps, LaTeX formatting, and assisting with parts of the experimental and
plotting scripts. We did not use generative AI tools to generate synthetic
datasets or modify experimental data, and all reported numbers come from actual
runs of open-weight models on public datasets. We reviewed and verified all
AI-assisted text, derivation checks, and code, checked all reported results
against raw experimental outputs, and take full responsibility for the final
content of this paper.

\subsection*{Reproducibility statement}

All experiments use public classification datasets and open-weight models,
with source classifiers from the Qwen3.5, Gemma, Llama, and Phi families and an
E5-small-v2 text reader; no proprietary models or restricted data are involved.
Section~\ref{sec:setup} specifies the tasks, models, and matched-panel protocol.
Appendix~\ref{app:edc-formal} states the formal definitions, execution
guarantees, proofs, and small-budget optimality bounds of EDC.
Appendix~\ref{app:rq1-protocol} documents the full frozen-interface construction
pipeline, including dataset partitions, prompt templates and vocabulary
exclusions, Jacobian lens calibration and readout, message-level indicator
construction, capacity-matched Surface baselines, executable rule fitting, and
reader training hyperparameters. Appendix~\ref{app:recovery-validation} reports
task-level statistical tests, error-following controls, cross-family panels, and
depth--budget fidelity grids. Appendix~\ref{app:interface-inspection} covers
non-literal activation analyses, rule composition, word-deletion interventions,
and alternative observation protocols, while
Appendix~\ref{app:reader-replacement} details the paired reader-replacement
evaluation and correction--harm decomposition. All reported metrics are computed
directly from held-out model and rule predictions, and the supplementary
materials provide the core J-Miner package, a runnable offline example, and
automated tests.

\subsection*{Ethics statement}

This study analyzes the internal decision logic of open-weight language-model
classifiers using established public benchmarks. No new human-subject data were
collected, no IRB approval was required, and there are no conflicts of interest
or external funding relationships to declare. Because benchmarks such as Civil
Comments and HateXplain contain offensive or toxic text, short excerpts and
vocabulary items appear in the paper solely to illustrate how classifiers process
such inputs. Making the decision rules of safety-related classifiers explicit
could in principle help adversaries probe decision boundaries. However, the rules
recovered here are checkpoint-specific, distribution-bounded approximations
rather than exact bypass recipes, and exposing them enables practitioners to
audit classifiers for spurious cues, non-literal triggers, and failure modes
before deployment.

\bibliographystyle{iclr2027_conference}
\bibliography{refs}

\clearpage
\appendix
\section{Formal Properties of Executable Decision Compression}
\label{app:edc-formal}

This appendix formalizes the Executable Decision Compression (EDC) framework introduced in Section~\ref{sec:edc}. We define the admissible representation class and prove single-variable screening optimality in Appendix~\ref{app:edc-admissible}, derive the margin preservation and decision-flip decompositions under reader replacement in Appendices~\ref{app:edc-guarantee-formal} and~\ref{app:edc-corrections}, and establish NP-hardness alongside exact small-budget bounds in Appendix~\ref{app:edc-small-budget}.

\subsection{Admissible Representations and Variable Ranking}
\label{app:edc-admissible}
\label{app:edc-objective-formal}

An observation protocol $\mathcal O$ specifies a finite set of candidate identifiers $\mathcal N_{\mathcal O}$ and an active event set $\mathcal E_{\mathcal O}(x)\subseteq\mathcal N_{\mathcal O}$ extracted from the source classifier's internal states $T_\theta(x)$ on input $x$. This defines the binary candidate evidence map
\begin{equation}
e_{\mathcal O}(x)=
\big(\mathbf 1[n\in\mathcal E_{\mathcal O}(x)]\big)_{n\in\mathcal N_{\mathcal O}}
\in\{0,1\}^{\mathcal N_{\mathcal O}}.
\label{eq:app-edc-evidence-map}
\end{equation}
J-Lens vocabulary readouts, vanilla logit-lens projections, and thresholded SAE activations instantiate $\mathcal O$ by extracting different candidate sets $\mathcal N_{\mathcal O}$ and active events $\mathcal E_{\mathcal O}(x)$ from $T_\theta(x)$, as detailed in Appendix~\ref{app:rq1-protocol}. All interfaces share the same downstream variable selection and rule-fitting formulation. For a depth cutoff $\ell$, the cumulative protocol $\mathcal O_{\ell_0:\ell}$ restricts readout aggregation to layers $\{\ell_0,\ldots,\ell\}$. Over $K$ selected binary variables, let $\mathcal G=\bigcup_{K\ge 1}\mathcal G_K$ denote a rule family mapping $\{0,1\}^K$ to $\mathcal Y$. Alongside bounded-size Boolean ASTs, we analyze the linear scorecard family
\begin{equation}
\mathcal G_{\mathrm{LR},K}
=\big\{c\mapsto\mathbf 1[b+w^\top c>0]:\
w\in\mathbb R^K,\ b\in\mathbb R\big\},
\label{eq:app-edc-rule-family}
\end{equation}
which extends to multiclass tasks via $c\mapsto\operatorname*{arg\,max}_{y\in\mathcal Y}(b_y+w_y^\top c)$ under a fixed tie-breaking rule.

\begin{definition}[Admissible representation and minimum state budget]
\label{def:edc-admissible}
\label{def:edc-objective}
A representation $\Pi=(\mathcal V,q_{\mathcal O,\mathcal V},g)$ belongs to the admissible class $\mathcal A(\mathcal O,\mathcal G)$ if and only if:
(i) $\mathcal V=(v_1,\ldots,v_K)$ consists of $K$ distinct identifiers from $\mathcal N_{\mathcal O}$;
(ii) $q_{\mathcal O,\mathcal V}(T_\theta(x))
=\big(\mathbf 1[v_j\in\mathcal E_{\mathcal O}(x)]\big)_{j=1}^K$ is the coordinate projection of $e_{\mathcal O}(x)$ onto $\mathcal V$; and
(iii) $g\in\mathcal G_K$.
For a tolerated disagreement $\delta\in[0,1]$, the minimum state budget $K_\delta^\star(\mathcal O,\mathcal G)$ in Equation~\eqref{eq:edc-objective} is the smallest $K=|\mathcal V|$ among $\Pi\in\mathcal A(\mathcal O,\mathcal G)$ satisfying $\varepsilon_\theta(\Pi)\le\delta$, and $+\infty$ if no such representation exists.
\end{definition}

Restricting $q_{\mathcal O,\mathcal V}$ to coordinate projection of $e_{\mathcal O}(x)$ separates the candidate evidence supplied by $\mathcal O$ from the complexity of the decision rule $g$. In Section~\ref{sec:edc-mining}, J-Miner selects $\mathcal V$ on the construction split $D_c$ by ranking support-filtered candidates according to their class-conditional trigger-rate gap $\Delta(v)=|\pi_1(v)-\pi_0(v)|$. Proposition~\ref{prop:edc-single-variable} shows that $\Delta(v)$ orders candidates by the highest class-balanced fidelity achievable by any single-variable binary rule on $D_c$.

\begin{proposition}[Single-variable ranking optimality]
\label{prop:edc-single-variable}
Fix an observation protocol $\mathcal O$. For any candidate $v\in\mathcal N_{\mathcal O}$, let $c_v(x)=\mathbf 1[v\in\mathcal E_{\mathcal O}(x)]$ and let $Y(x)=f_\theta(x)\in\{0,1\}$ denote the source classifier's prediction on $D_c$, assuming both predicted classes are nonempty. Write $\pi_y(v)=\widehat{\Pr}_{D_c}(c_v=1\mid Y=y)$ for $y\in\{0,1\}$, and define the empirical class-balanced fidelity of a deterministic single-variable rule $h:\{0,1\}\to\{0,1\}$ on $D_c$ as
\begin{equation}
\widehat{\operatorname{BFid}}_{D_c}(h)
=\frac12\widehat{\Pr}_{D_c}\big[h(c_v)=1\mid Y=1\big]
+\frac12\widehat{\Pr}_{D_c}\big[h(c_v)=0\mid Y=0\big].
\label{eq:app-edc-bfid}
\end{equation}
Then the maximum class-balanced fidelity over all deterministic rules on $c_v$ is
\begin{equation}
\max_{h:\{0,1\}\to\{0,1\}}
\widehat{\operatorname{BFid}}_{D_c}(h)
=\frac12+\frac12\Delta(v),
\qquad
\Delta(v)=|\pi_1(v)-\pi_0(v)|.
\label{eq:app-edc-single-opt}
\end{equation}
\end{proposition}

\begin{proof}
The four functions from $\{0,1\}$ to $\{0,1\}$ are the constant rules $h\equiv 0$ and $h\equiv 1$, the identity rule $h(c_v)=c_v$, and the negation rule $h(c_v)=1-c_v$. Direct substitution into Equation~\eqref{eq:app-edc-bfid} yields balanced fidelity $1/2$ for both constant rules, $(1+\pi_1(v)-\pi_0(v))/2$ for the identity rule, and $(1-\pi_1(v)+\pi_0(v))/2$ for the negation rule. Taking the maximum over the four rules gives $1/2+|\pi_1(v)-\pi_0(v)|/2$.
\end{proof}

Selecting the top-$K$ candidates under $\Delta(v)$ forms an ordered basis $\mathcal V=(v_1,\ldots,v_K)$ over which the linear scorecard or Boolean AST captures multi-variable combinations. Appendix~\ref{app:edc-small-budget} compares J-Miner's fitted linear scorecards with construction-optimal linear rules at small budgets.

\subsection{Execution Guarantee for a Fixed Rule}
\label{app:edc-guarantee-formal}

In standalone execution (Section~\ref{sec:edc-transfer}), an independent text reader replaces the source evidence state $c_\theta(x)$ with a predicted state $\widetilde c_\phi(x)$, and coordinate mismatches need not alter the output of the frozen rule $g$. Consider a binary scorecard with score $z(c)=b+w^\top c$, decision $g(c)=\mathbf 1[z(c)>0]$, margin $\gamma(c)=|z(c)|$, and decision sign $\sigma(c)=2g(c)-1\in\{-1,+1\}$. Replacing $c\in\{0,1\}^K$ with $\widetilde c\in\{0,1\}^K$ shifts coordinate $j$'s contribution by $\delta_j=w_j(\widetilde c_j-c_j)$. Writing $(u)_+=\max(u,0)$, we decompose this shift into adverse and favorable perturbations
\begin{equation}
E_w^-(c,\widetilde c)=\sum_{j=1}^K\big(-\sigma(c)\,\delta_j\big)_+,\qquad
E_w^+(c,\widetilde c)=\sum_{j=1}^K\big(\sigma(c)\,\delta_j\big)_+.
\label{eq:app-edc-signed-mass}
\end{equation}
Here $E_w^-(c,\widetilde c)$ sums the weighted coordinate changes that push the score toward the decision boundary, while $E_w^+(c,\widetilde c)$ sums the changes that push the score further in the direction of $\sigma(c)$.

\begin{proposition}[Decision preservation]
\label{prop:edc-preservation-formal}
For any binary states $c,\widetilde c\in\{0,1\}^K$, the following properties hold:
(a) $\sigma(c)\,z(\widetilde c)=\gamma(c)+E_w^+(c,\widetilde c)
-E_w^-(c,\widetilde c)$.
(b) If $g(c)=1$, then $g(\widetilde c)=g(c)$ if and only if $E_w^-(c,\widetilde c)<\gamma(c)+E_w^+(c,\widetilde c)$. If $g(c)=0$, then $g(\widetilde c)=g(c)$ if and only if $E_w^-(c,\widetilde c)\le\gamma(c)+E_w^+(c,\widetilde c)$.
(c) If $E_w^-(c,\widetilde c)<\gamma(c)$, then $g(\widetilde c)=g(c)$.
\end{proposition}

\begin{proof}
If $g(c)=1$, then $z(c)>0$; if $g(c)=0$, then $z(c)\le0$. In both cases $\sigma(c)z(c)=\gamma(c)$. Since $z(\widetilde c)=z(c)+\sum_j\delta_j$ and $\sigma(c)\delta_j=(\sigma(c)\delta_j)_+-(-\sigma(c)\delta_j)_+$ for each $j$, multiplying by $\sigma(c)$ gives (a). If $g(c)=1$, the decision is preserved if and only if $z(\widetilde c)>0$, that is, $\sigma(c)z(\widetilde c)>0$. If $g(c)=0$, it is preserved if and only if $z(\widetilde c)\le0$, that is, $\sigma(c)z(\widetilde c)\ge0$. Substituting (a) gives (b). Under condition (c), part (a) gives $\sigma(c)z(\widetilde c)>E_w^+(c,\widetilde c)\ge0$, satisfying both cases of (b).
\end{proof}

On the test set, Table~\ref{tab:transfer} reports Cover as the fraction of paired states $c=c_\theta(x)$ and $\widetilde c=\widetilde c_\phi(x)$ with $E_w^-(c,\widetilde c)<\gamma(c)$, and Keep as the fraction with $g(\widetilde c)=g(c)$. Proposition~\ref{prop:edc-preservation-formal}(c) implies $\mathrm{Cover}\le\mathrm{Keep}$; the difference $\mathrm{Keep}-\mathrm{Cover}$ reflects inputs where $E_w^+(c,\widetilde c)$ offsets $E_w^-(c,\widetilde c)\ge\gamma(c)$.

\subsection{Corrections and Harms}
\label{app:edc-corrections}

To analyze how the remaining decision flips affect agreement with the source classifier in Section~\ref{sec:edc-transfer} and \hyperref[sec:rq3-transfer]{RQ3}, let $Y=f_\theta(X)\in\{0,1\}$ denote the source classifier's verdict, $\widehat Y_{\mathrm s}=g(c_\theta(X))$ the source-state rule decision with fidelity $F_\Pi=\Pr[\widehat Y_{\mathrm s}=Y]=1-\varepsilon_\theta(\Pi)$, and $\widehat Y_\phi=g(\widetilde c_\phi(X))$ the standalone decision after reader replacement with fidelity $F_\phi=\Pr[\widehat Y_\phi=Y]$. We partition all decision changes across the input distribution into corrections and harms with probabilities
\begin{equation}
p_{\mathrm{corr}}
=\Pr\big[\widehat Y_{\mathrm s}\neq Y,\ \widehat Y_\phi=Y\big],
\qquad
p_{\mathrm{harm}}
=\Pr\big[\widehat Y_{\mathrm s}=Y,\ \widehat Y_\phi\neq Y\big].
\label{eq:app-edc-corr-harm}
\end{equation}
Because $Y,\widehat Y_{\mathrm s},\widehat Y_\phi\in\{0,1\}$, every binary decision flip $\{\widehat Y_\phi\neq\widehat Y_{\mathrm s}\}$ either resolves a source-state disagreement with $Y$ or introduces a new one. These two events are disjoint, so their sum equals the total decision-flip rate $1-\mathrm{Keep}=\Pr[\widehat Y_\phi\neq\widehat Y_{\mathrm s}]$. Subtracting $F_\Pi=\Pr[\widehat Y_{\mathrm s}=Y,\widehat Y_\phi=Y]+p_{\mathrm{harm}}$ from $F_\phi=\Pr[\widehat Y_{\mathrm s}=Y,\widehat Y_\phi=Y]+p_{\mathrm{corr}}$ yields the exact net fidelity change $F_\phi-F_\Pi=p_{\mathrm{corr}}-p_{\mathrm{harm}}$.

Whether a state perturbation induces a correction or a harm is determined by the verdict-aligned score shift $\Delta_Y=(2Y-1)\big(z(\widetilde c)-z(c)\big)$ for $c=c_\theta(X)$ and $\widetilde c=\widetilde c_\phi(X)$, which is positive whenever replacement moves the rule score toward the class predicted by $f_\theta$.

\begin{proposition}[Decision-change decomposition]
\label{prop:edc-net-change}
\label{prop:edc-direction}
If $z(\widetilde c)\neq0$ almost surely, then corrections and harms take the threshold forms
\begin{equation}
p_{\mathrm{corr}}=\Pr\big[\widehat Y_{\mathrm s}\neq Y,\ \Delta_Y>\gamma(c)\big],
\qquad
p_{\mathrm{harm}}=\Pr\big[\widehat Y_{\mathrm s}=Y,\ \Delta_Y<-\gamma(c)\big],
\label{eq:app-edc-direction}
\end{equation}
and the net change in source-classifier fidelity satisfies Equation~\eqref{eq:edc-decomposition}:
\[
F_\phi-F_\Pi
=\Pr\big[g(c)\neq Y,\ \Delta_Y>\gamma(c)\big]
-\Pr\big[g(c)=Y,\ \Delta_Y<-\gamma(c)\big].
\]
\end{proposition}

\begin{proof}
Let $\tau=2Y-1\in\{-1,+1\}$. When $\widehat Y_{\mathrm s}=Y$, the sign of $z(c)$ matches $\tau$, so $\tau z(c)=|z(c)|=\gamma(c)$ and $\tau z(\widetilde c)=\gamma(c)+\Delta_Y$. Since $z(\widetilde c)\neq0$, the replaced decision $\widehat Y_\phi$ disagrees with $Y$ if and only if $\tau z(\widetilde c)<0$, which is equivalent to $\Delta_Y<-\gamma(c)$. Conversely, when $\widehat Y_{\mathrm s}\neq Y$, $\tau z(c)=-\gamma(c)$ and $\tau z(\widetilde c)=-\gamma(c)+\Delta_Y$, so $\widehat Y_\phi$ agrees with $Y$ if and only if $\tau z(\widetilde c)>0$, or equivalently $\Delta_Y>\gamma(c)$. Substituting these two characterizations into Equation~\eqref{eq:app-edc-corr-harm} and taking their difference $F_\phi-F_\Pi=p_{\mathrm{corr}}-p_{\mathrm{harm}}$ yields Equation~\eqref{eq:edc-decomposition}.
\end{proof}

Proposition~\ref{prop:edc-direction} shows why replacing $c_\theta(x)$ with $\widetilde c_\phi(x)$ can raise source-classifier fidelity: any positive shift $\Delta_Y>0$ preserves agreeing decisions and converts disagreeing decisions with $\gamma(c)<\Delta_Y$ into corrections, as quantified across all six binary tasks in Appendix~\ref{app:reader-replacement}.

\subsection{Exact Optima and Budget Bounds at Small State Sizes}
\label{app:edc-small-budget}

Any admissible $K$-variable linear rule achieving distortion at most $\delta$ on the empirical construction distribution $\widehat{\mathcal D}_c$ provides an upper bound $K_\delta^\star(\mathcal O,\mathcal G_{\mathrm{LR}})\le K$. To certify matching lower bounds, we first show that computing $K_\delta^\star$ is NP-hard in general and then combine exhaustive small-budget evaluation with a support-filtering bound on $\widehat{\mathcal D}_c$.

\begin{proposition}[Hardness of the ideal budget]
\label{prop:edc-hardness}
When a finite input distribution, its predicted labels, and a candidate event map are given as input, computing $K_0^\star(\mathcal O,\mathcal G_{\mathrm{LR}})$ is NP-hard.
\end{proposition}

\begin{proof}
We reduce from set cover \citep{karp1972reducibility}, following standard reductions for minimum feature subsets \citep{davies1994np,amaldi1998approximability}. Given subsets $S_1,\ldots,S_n$ of $\{1,\ldots,m\}$, let the distribution be uniform over inputs $x_1,\ldots,x_m$ with label $1$ and one input $x_0$ with label $0$. Candidate $v_i$ is active on $x_e$ exactly when $e\in S_i$ and inactive on $x_0$. A cover $T$ yields the zero-distortion rule $g(c)=\mathbf 1[\sum_{i\in T}c_i-1/2>0]$. Conversely, if a rule over variables $T$ has zero distortion, each $x_e$ activates some variable in $T$, since its state would otherwise equal that of $x_0$; hence $T$ is a cover. Thus $K_0^\star$ equals the minimum cover size, establishing NP-hardness.
\end{proof}

Because the reduction in Proposition~\ref{prop:edc-hardness} uses a disjunction of positive literals, the same intractability holds for Boolean ASTs whose size bound admits a disjunction over the selected variables. At small budgets $k\in\{1,2\}$, the maximum construction-split fidelity $F^{\max}_k$ of an admissible $k$-variable linear representation can be evaluated exhaustively. A rule on $k$ binary variables assigns a binary prediction to each of the $2^k$ joint states, so its fidelity on $\widehat{\mathcal D}_c$ depends only on the class-conditional counts of those joint states. Every Boolean labeling on one or two variables is realizable by a linear threshold rule in $\mathcal G_{\mathrm{LR},k}$ except XOR and XNOR at $k=2$, where the optimal linear threshold rule flips the single joint state with the smallest class-count difference. Because exhaustive search runs over the support-filtered candidate pool from Appendix~\ref{app:rq1-protocol}, Proposition~\ref{prop:edc-support-restriction} bounds the maximum fidelity contribution of omitted low-support candidates to certify bounds over $\mathcal N_{\mathcal O}$.

\begin{proposition}[Support restriction]
\label{prop:edc-support-restriction}
Adding one variable whose construction support rate is $\rho$ to a variable set raises its best construction fidelity by at most $\rho$. Consequently, if a candidate pool omits only names with support rate at most $\rho_{\mathrm{ex}}$,
\begin{equation}
F^{\max}_k\le\max_{0\le j\le k}
\big\{F^{\max}_{k-j}(\mathrm{pool})+j\,\rho_{\mathrm{ex}}\big\},
\label{eq:app-edc-support-restriction}
\end{equation}
where $F^{\max}_0$ is the majority-class rate.
\end{proposition}

\begin{proof}
Given a linear rule on $S\cup\{v\}$, setting $c_v=0$ gives a linear rule on $S$ that agrees with it wherever $v$ is inactive. The two rules can disagree only on inputs where $v$ is active, which have probability $\rho$. Applying this to each omitted variable gives the bound.
\end{proof}

Combining exhaustive evaluation of $F^{\max}_1(\mathrm{pool})$ and $F^{\max}_2(\mathrm{pool})$ with Proposition~\ref{prop:edc-support-restriction} under the 3\% support filter $\rho_{\mathrm{ex}}<0.03$ yields two-sided bounds on $K_\delta^\star$ over $\widehat{\mathcal D}_c$. If the bound in Equation~\eqref{eq:app-edc-support-restriction} falls below $1-\delta$ for $k\in\{1,2\}$, no admissible $k$-variable rule over $\mathcal N_{\mathcal O}$ achieves distortion $\delta$, certifying $K_\delta^\star>k$. Conversely, any admissible representation whose construction-split fidelity reaches $1-\delta$ certifies $K_\delta^\star\le K$, determining $K_\delta^\star$ whenever both bounds meet.

Table~\ref{tab:edc-small-budget} compares construction-optimal linear rules selected within the 3\% support-filtered candidate pool with J-Miner's ranked-prefix scorecards across all six binary tasks, reporting $F^{\max}_k(\mathrm{pool})$ under Constr., held-out fidelity of that fixed rule under Test, and J-Miner's held-out fidelity under Prefix test. J-Miner's internal rules reach mean held-out fidelity of 76.7\% at $K=1$ and 81.1\% at $K=2$, compared with 76.8\% and 82.1\% for the construction-optimal internal rules, a held-out fidelity gap of 1.0~percentage point at $K=2$. Construction-optimal literal-word rules reach mean test fidelities of 62.5\% and 65.8\%, respectively. On every task at both budgets, internal variables yield higher construction optima and higher test fidelity for the selected rules than literal words. Figure~\ref{fig:compression-evidence}a reports budget curves for the five-task subset with at least 32 support-filtered surface words, and Appendix~\ref{app:parameter-sensitivity} compares ranked prefixes with random orderings at larger budgets.

At $\delta=0.10$, corresponding to 90\% fidelity on $\widehat{\mathcal D}_c$, the bounds in Table~\ref{tab:edc-small-budget} account for omitted low-support candidates via Proposition~\ref{prop:edc-support-restriction} and certify exact minimum budgets of $K_{0.10}^\star=1$ for SMS, $K_{0.10}^\star=2$ for Toxicity, and $K_{0.10}^\star=3$ for Formality, each achieved by J-Miner's ranked prefix, whereas $\ge 3$ indicates $K_{0.10}^\star>2$ for all literal surface-word pools and the remaining tasks.

\begin{table}[!ht]
  \centering
  \caption{Small-budget linear-rule fidelity (\%) and certified minimum budgets ($K^\star_{0.10}$) across the six binary tasks.}
  \label{tab:edc-small-budget}
  \smallskip
  \small
  \renewcommand{\arraystretch}{0.90}
  \setlength{\aboverulesep}{0.4ex}
  \setlength{\belowrulesep}{0.5ex}
  \setlength{\tabcolsep}{4pt}
  \begin{tabular}{llrrrrrrc}
    \toprule
    & & \multicolumn{2}{c}{Exhaustive $K{=}1$} & \multicolumn{2}{c}{Exhaustive $K{=}2$}
    & \multicolumn{2}{c}{Prefix test} & \\
    \cmidrule(lr){3-4}\cmidrule(lr){5-6}\cmidrule(lr){7-8}
    Task & Evidence & Constr. & Test & Constr. & Test & $K{=}1$ & $K{=}2$
    & $K^\star_{0.10}$ \\
    \midrule
    SMS       & J-Lens & 96.7 & 95.0 & 97.7 & 96.7 & 95.0 & 96.7 & 1 \\
              & Literal & 69.3 & 73.3 & 79.1 & 81.7 & 73.3 & 80.7 & $\ge3$ \\
    Sentiment & J-Lens & 72.2 & 70.7 & 78.8 & 77.5 & 69.8 & 75.8 & $\ge3$ \\
              & Literal & 55.5 & 55.3 & 57.8 & 58.3 & 50.0 & 52.2 & $\ge3$ \\
    Formality & J-Lens & 85.3 & 82.8 & 89.8 & 87.3 & 83.3 & 85.3 & 3 \\
              & Literal & 73.1 & 72.8 & 74.6 & 75.7 & 72.8 & 76.5 & $\ge3$ \\
    IMDB      & J-Lens & 77.7 & 76.0 & 82.3 & 81.2 & 76.0 & 79.0 & $\ge3$ \\
              & Literal & 60.4 & 63.5 & 64.0 & 66.8 & 63.5 & 63.5 & $\ge3$ \\
    Toxicity  & J-Lens & 83.2 & 77.8 & 90.5 & 86.3 & 77.8 & 86.0 & 2 \\
              & Literal & 53.1 & 52.0 & 55.3 & 52.2 & 47.2 & 54.3 & $\ge3$ \\
    Sarcasm   & J-Lens & 62.9 & 58.3 & 67.6 & 63.7 & 58.3 & 63.7 & $\ge3$ \\
              & Literal & 59.8 & 57.8 & 63.2 & 60.3 & 57.8 & 60.3 & $\ge3$ \\
    \midrule
    Mean      & J-Lens & 79.7 & 76.8 & 84.5 & 82.1 & 76.7 & 81.1 & \\
              & Literal & 61.9 & 62.5 & 65.7 & 65.8 & 60.8 & 64.6 & \\
    \bottomrule
  \end{tabular}
\end{table}
\FloatBarrier

\section{Frozen Concept--Rule Construction and Reader Training}
\label{app:rq1-protocol}

This appendix details dataset partitioning and source-classifier training
(Appendix~\ref{app:data-separation}), J-Lens readout pooling and variable
selection (Appendix~\ref{app:lens-readout}), rule fitting
(Appendix~\ref{app:rule-fitting}), and standalone text-reader training
(Appendix~\ref{app:student-reconstruction}).

\subsection{Data Separation and Target-Model Freezing}
\label{app:data-separation}

Table~\ref{tab:classifier-config} lists the disjoint training, validation, and
test partitions subsampled from standard public splits
\citep{almeida2011contributions,barbieri2020tweeteval,pavlick2016empirical,maas2011learning,borkan2019nuanced,misra2019sarcasm,mathew2021hatexplain,coucke2018snips},
with Civil Comments toxicity binarized at $0.5$. All binary test sets are
class-balanced. The construction split $D_c$ combines training and validation
examples except for SNIPS-3, which uses its 900 validation examples. Candidate
support filtering, variable ranking, and rule fitting use $D_c$ and the frozen
source predictions $y_\theta(x)=f_\theta(x)$, with all selected variables and
rules fixed prior to test evaluation.

We fine-tune a separate Qwen3.5-0.8B classifier on each task's training split
using single-token decision prompts. Every binary prompt follows the template
``You are a binary \texttt{<task phrase>} classifier. Return exactly one label
token. Labels: 1 = \texttt{<label 1>}, 0 = \texttt{<label 0>}.
\texttt{<Field>}: \texttt{<text>} \texttt{Decision:}'' with the fields in
Table~\ref{tab:prompt-exclusion}. We apply LoRA adapters with rank $16$,
$\alpha=32$, and dropout $0.05$ to all attention and MLP projection matrices and
train them with AdamW at learning rate $2\times10^{-4}$ and weight decay $0.01$
under the epoch, microbatch, and gradient-accumulation schedules in
Table~\ref{tab:classifier-config}. Training minimizes next-token cross-entropy
at the terminal prompt position against the gold label token. We select the
checkpoint with the highest validation macro-F1 among those with valid
single-token label emission, merge the adapters, and freeze the classifier
$f_\theta$.

For the reference baselines in Table~\ref{tab:recovery}, Zero-shot uses the
unadapted Qwen3.5-0.8B backbone with native chat formatting and thinking
disabled, taking the argmax over valid label-token logits. Majority predicts the
most frequent source-classifier label on $D_c$. Paired metric differences use
$10{,}000$ bootstrap resamples of held-out examples for confidence intervals,
and paired classifier comparisons use exact two-sided McNemar tests.

\begin{table}[htbp]
  \centering
  \caption{Dataset partitions, construction sample sizes ($|D_c|$), and
  fine-tuning schedules for Qwen3.5-0.8B source classifiers.}
  \label{tab:classifier-config}
  \medskip
  \small
  \setlength{\tabcolsep}{4pt}
  \begin{tabular*}{\linewidth}{@{\extracolsep{\fill}}lrrrrr@{}}
    \toprule
    Task & Train / Val. / Test & $|D_c|$ & Epochs & Batch $\times$ Accum. & Max Tokens \\
    \midrule
    SMS        & 896 / 298 / 300     & 1,194 & 4 & $8\times2$ & 192 \\
    Sentiment  & 1,800 / 600 / 600   & 2,400 & 4 & $8\times2$ & 256 \\
    Formality  & 1,800 / 600 / 600   & 2,400 & 4 & $8\times2$ & 256 \\
    IMDB       & 1,800 / 600 / 600   & 2,400 & 4 & $4\times4$ & 512 \\
    Toxicity   & 1,800 / 600 / 600   & 2,400 & 4 & $8\times2$ & 256 \\
    Sarcasm    & 1,800 / 600 / 600   & 2,400 & 4 & $8\times2$ & 256 \\
    HateXplain & 1,800 / 600 / 600   & 2,400 & 4 & $8\times4$ & 512 \\
    SNIPS-3    & 4,800 / 900 / 297   & 900   & 3 & $8\times4$ & 128 \\
    SNIPS-7    & 4,200 / 1,400 / 665 & 5,600 & 4 & $8\times4$ & 256 \\
    \bottomrule
  \end{tabular*}
\end{table}

\subsection{From Internal Readouts to Message-Level Variables}
\label{app:lens-readout}
\label{app:indicator-construction}

The observation protocol in Section~\ref{sec:edc-mining} uses J-Lens
\citep{gurnee2026workspace} to map intermediate residual states into the final
residual space. Writing $h_{\ell,p}(x)$ for the state at layer $\ell$ and token
position $p$, we estimate the transport operator to the final layer $L$ as
\begin{equation}
J_\ell=\mathbb{E}_{(x,p)}\left[
  \sum_{p'\geq p}
  \frac{\partial h_{L,p'}(x)}{\partial h_{\ell,p}(x)}
\right].
\label{eq:jacobian}
\end{equation}
We estimate this expectation on 1,000 WikiText-103 training passages of up to
128 tokens each \citep{merity2017pointer}, independently of the classification
datasets, averaging over interior token positions and across passages after
excluding the first 16 tokens and the final token.
Applying final normalization $\mathrm{Norm}$ and vocabulary unembedding matrix
$U$ gives the readout scores
\begin{equation}
r_{\ell,p}(x)=U\,\mathrm{Norm}\!\left(J_\ell h_{\ell,p}(x)\right),
\label{eq:vocabulary-readout}
\end{equation}
which rank vocabulary coordinates at each inspected layer and position.

We pool these readouts over the input's content span, excluding the instruction
template and terminal \texttt{Decision:} position. The layer windows in
Table~\ref{tab:concept-layer-windows} are fixed before held-out evaluation.
For binary classifiers, the transported layers are supplemented by a final-layer
identity readout $U\,\mathrm{Norm}(h_{L,p}(x))$; the multiclass protocol uses
only transported layers.

At every inspected layer and content position, we retain the top $R=3$
vocabulary coordinates of $r_{\ell,p}(x)$, strip and lowercase their decoded
strings, and discard fragments shorter than two characters. By exact matching
after normalization, we exclude the task-specific terms in
Table~\ref{tab:prompt-exclusion} together with 18 shared template words:
\texttt{0}, \texttt{1}, \texttt{yes}, \texttt{no}, \texttt{true},
\texttt{false}, \texttt{zero}, \texttt{one}, \texttt{binary},
\texttt{classifier}, \texttt{decision}, \texttt{exactly}, \texttt{label},
\texttt{labels}, \texttt{return}, \texttt{token}, \texttt{positive}, and
\texttt{negative}. Taking the union of the remaining
normalized names across the content span and layer window forms the
message-level event set $\mathcal{E}_{\mathcal{O}}(x)$, so that occurrences of
the same normalized token at different positions or depths merge into a single
binary indicator:
\begin{equation}
  c_v(x)=\mathbf{1}\big[v\in\mathcal{E}_{\mathcal{O}}(x)\big],
  \qquad
  c_{\theta,j}(x)=c_{v_j}(x),\quad j=1,\ldots,K.
  \label{eq:app-message-indicator}
\end{equation}

\begin{table}[htbp]
  \centering
  \caption{Fixed zero-based layer windows for message-level variable extraction.}
  \label{tab:concept-layer-windows}
  \medskip
  \small
  \begin{tabular*}{\linewidth}{@{\extracolsep{\fill}}llcc@{}}
    \toprule
    Classifier & Tasks & Transported layers & Identity \\
    \midrule
    Qwen3.5-0.8B & Binary     & 14--22 & 23 \\
    Qwen3.5-2B   & Binary     & 14--22 & 23 \\
    Qwen3.5-4B   & Binary     & 19--30 & 31 \\
    Gemma-3-1B   & Binary     & 16--24 & 25 \\
    Llama-3.2-1B & Binary     & 10--14 & 15 \\
    Phi-4-mini   & Binary     & 20--30 & 31 \\
    Qwen3.5-0.8B & Multiclass & 17--22 & None \\
    \bottomrule
  \end{tabular*}
\end{table}

\begin{table}[htbp]
  \centering
  \caption{Prompt template fields and task-specific excluded vocabulary for the
  six binary classification tasks.}
  \label{tab:prompt-exclusion}
  \medskip
  \small
  \setlength{\tabcolsep}{3pt}
  \begin{tabularx}{\linewidth}{@{}llllX@{}}
    \toprule
    Task & Task phrase & Labels (1 / 0) & Field & Additional excluded names \\
    \midrule
    SMS       & SMS spam  & spam / ham                & Message  & spam, ham, sms, message, classification \\
    Sentiment & sentiment & positive / negative       & Text     & sentiment, text \\
    Formality & formality & formal / informal         & Sentence & formal, informal, formality, sentence \\
    IMDB      & sentiment & positive / negative       & Review   & sentiment, review \\
    Toxicity  & toxicity  & toxic / non-toxic         & Comment  & toxic, toxicity, non, comment, classification \\
    Sarcasm   & sarcasm   & sarcastic / not sarcastic & Headline & sarcastic, sarcasm, not, headline \\
    \bottomrule
  \end{tabularx}
\end{table}

We then select the compact concept basis $\mathcal{V}$ on $D_c$ against the
frozen source predictions $y_\theta(x)=f_\theta(x)$. Let $n_y$ denote the number
of construction examples assigned class $y\in\mathcal{Y}$. For each candidate
variable $v$, we compute its empirical support and class-conditional trigger
rate on $D_c$:
\begin{equation}
\operatorname{supp}(v)=\frac{1}{|D_c|}\sum_{x\in D_c}c_v(x),\qquad
\pi_y(v)=\frac{1}{n_y}
\sum_{\substack{x\in D_c\\y_\theta(x)=y}}c_v(x).
\label{eq:concept-statistics}
\end{equation}
We discard low-frequency candidates with $\operatorname{supp}(v)<0.03$ and rank
the remaining candidates by their maximum one-vs-rest trigger-rate gap
$\Delta(v)=\max_{y\in\mathcal{Y}}|\pi_y(v)-\pi_{\neg y}(v)|$, where
$\pi_{\neg y}(v)$ is the trigger rate across construction examples assigned to
classes other than $y$. For binary classification, this statistic reduces to
$\Delta(v)=|\pi_1(v)-\pi_0(v)|$; by
Proposition~\ref{prop:edc-single-variable}, $\Delta(v)$ ranks candidates by
their optimal single-variable class-balanced fidelity to $f_\theta$ on $D_c$.
Taking the top $K=16$ candidates in decreasing order of $\Delta(v)$ yields the
ordered variable set $\mathcal{V}=(v_1,\ldots,v_K)$.

\label{app:surface-construction}
For the matched linear-scorecard comparisons, the Surface baseline replaces the
internal event set $\mathcal{E}_{\mathcal{O}}(x)$ in
Equation~\eqref{eq:app-message-indicator} with the literal words $\tau(x)$ in
the raw input. We lowercase the text, tokenize at word boundaries, and retain
words of at least two characters. On the same split $D_c$ and source
predictions, we apply the $3\%$ support threshold and rank candidates by
$\Delta(v)$, taking the top $K$ words to form $c_S(x)\in\{0,1\}^K$. Both routes
share the variable budget and linear objective below.

\subsection{Executable Rule Fitting}
\label{app:rule-fitting}

We fit Boolean trees and linear scorecards on $D_c$ over the $K$-dimensional
binary state vector $c_\theta(x)\in\{0,1\}^K$. For the bounded Boolean AST rules
at $K=16$ in Table~\ref{tab:recovery}, we search over logical trees formed by
\textsc{and} and \textsc{or} operators applied to positive or negated concept
literals $c_{\theta,j}(x)$ and $1-c_{\theta,j}(x)$. Each candidate tree uses at
most five leaves and depth at most three, with each base variable appearing at
most once, and expressions are canonicalized under associativity and
commutativity to eliminate algebraically equivalent trees. On the six binary
tasks, we select the canonical tree that maximizes source-classifier fidelity on
$D_c$. On multiclass tasks, we fit one tree per class $y\in\mathcal{Y}$ to
maximize one-vs-rest F1 against $\mathbf{1}[y_\theta(x)=y]$ within the same
grammar. For a class-$y$ tree output $a\in\{0,1\}$, the Laplace-smoothed score
is $s_y(a)=(n_{y,a}+1)/(m_{y,a}+|\mathcal{Y}|)$, where $m_{y,a}$ counts
construction inputs producing output $a$ and $n_{y,a}$ counts those also
assigned class $y$ by $f_\theta$. We predict the class with the highest
calibrated score, breaking ties toward the lower class index.

For the linear scorecard $z(c)=b+w^\top c$ in
Equation~\eqref{eq:edc-scorecard}, we fit the signed weights $w\in\mathbb{R}^K$
and intercept $b\in\mathbb{R}$ to the source classifier's predictions on $D_c$.
For binary tasks, writing $\widetilde{y}_x=2y_\theta(x)-1\in\{-1,+1\}$ and
inverse class-frequency weights $\alpha_y=|D_c|/(2n_y)$, the scorecard
parameters solve the class-balanced $\ell_2$-regularized logistic objective
\begin{equation}
(w^*,b^*)=\operatorname*{arg\,min}_{w\in\mathbb{R}^K,\,b\in\mathbb{R}}
\sum_{x\in D_c}\alpha_{y_\theta(x)}
\log\!\left(1+\exp\!\big(-\widetilde{y}_x(b+w^\top c_\theta(x))\big)\right)
+\frac{\lambda}{2}\lVert w\rVert_2^2,
\label{eq:scorecard-objective}
\end{equation}
with unit regularization penalty $\lambda=1$. Binary predictions are positive
when $z(c_\theta(x))>0$. For multiclass tasks, we fit a class-balanced
multinomial logistic scorecard with the same unit $\ell_2$ penalty $\lambda=1$
and predict $\operatorname*{arg\,max}_{y\in\mathcal{Y}}(b_y+w_y^\top c_\theta(x))$.

\subsection{Reader State Reconstruction}
\label{app:student-reconstruction}

In Section~\ref{sec:rq3-transfer} and Table~\ref{tab:transfer}, we keep the
16-variable vocabulary $\mathcal{V}$ and scorecard $g$ fixed and train a
33.2M-parameter E5-small-v2 text reader $s_\phi:\mathcal{X}\to[0,1]^K$ on the
six binary tasks to predict the source concept state $c_\theta(x)$ from raw
input text. For each training example, running J-Lens and the indicator map of
Appendix~\ref{app:indicator-construction} provides the binary target
$c_\theta(x)$, while the frozen source classifier's final label-token scores
supply the two-class logit vector $a_\theta(x)$ whose argmax defines
$f_\theta(x)$. Let $n_j^+$ and $n_j^-$ denote the positive and negative counts
of coordinate $j$ in the reader-training split, with positive-class weight
$\rho_j=\max(n_j^-,1)/\max(n_j^+,1)$. The J-only reader minimizes the weighted
binary cross-entropy
\begin{equation}
\begin{aligned}
\mathcal{L}_{\mathrm{concept}}(x)
=-\frac{1}{K}\sum_{j=1}^{K}\big[
  &\rho_jc_{\theta,j}(x)\log s_{\phi,j}(x)\\
  &+(1-c_{\theta,j}(x))\log(1-s_{\phi,j}(x))\big].
\end{aligned}
\label{eq:student-concept-loss}
\end{equation}
The J+Aux reader attaches a training-only binary head $u_\phi$ to the shared
E5-small-v2 encoder to align the encoder representation with the source
classifier's soft verdict distribution. With temperature-scaled distributions
$p_{\theta,T}(x)=\operatorname{softmax}(a_\theta(x)/T)$ and
$p_{\mathrm{aux},T}(x)=\operatorname{softmax}([0,u_\phi(x)]/T)$ in class order
$(0,1)$ and weight $\beta=0.5$, J+Aux optimizes the joint objective
\begin{equation}
\mathcal{L}(x)=\mathcal{L}_{\mathrm{concept}}(x)
+\beta T^2\operatorname{KL}\big(p_{\theta,T}(x)\,\|\,p_{\mathrm{aux},T}(x)\big).
\label{eq:student-joint-loss}
\end{equation}

At inference, J+Aux discards its auxiliary head. Both readers binarize the
predicted concept probabilities at $0.5$ via
$\widetilde{c}_\phi(x)=\mathbf{1}[s_\phi(x)\ge 0.5]$ and evaluate the frozen
scorecard $\widehat{f}_\phi(x)=g(\widetilde{c}_\phi(x))$ in
Equation~\eqref{eq:edc-student}. As an unconstrained distillation baseline in
Table~\ref{tab:transfer}, Plain KD pairs the same E5-small-v2 encoder with a
direct classification head $v_\phi(x)$, minimizing
$T^2\operatorname{KL}\big(p_{\theta,T}(x)\,\|\,\operatorname{softmax}([0,v_\phi(x)]/T)\big)$
against the same teacher distributions $p_{\theta,T}(x)$ and predicting
$\mathbf{1}[\sigma(v_\phi(x))\ge 0.5]$ without intermediate concept states or a
rule layer.

All reader variants share the dataset boundaries in
Table~\ref{tab:classifier-config}, partitioning each task's construction split
$D_c$ into an $85\%/15\%$ reader-training and inner-development split while
evaluating on the identical held-out test set. Every model is trained for up to
three epochs with batch size 32 using AdamW and a projection-head learning rate
of $10^{-3}$. Across the three epochs, J-only evaluates encoder learning rates
$\{10^{-5}, 2\times10^{-5}, 4\times10^{-5}\}$, whereas J+Aux and Plain KD fix
the encoder learning rate at $2\times10^{-5}$ and evaluate softmax temperatures
$T\in\{1,2,4\}$. Across all three methods, the deployed checkpoint is selected
on the inner-development split by maximizing agreement with $f_\theta$ along
each method's inference path. Appendix~\ref{app:reader-replacement} reports the
resulting task-level decision changes.

\section{Extended Validation of Executable Recovery}
\label{app:recovery-validation}

This appendix supplements Section~\ref{sec:rq1-recover} with task-level
statistical comparisons, error-following results, and full evaluations across
model families, feature budgets, and observation depths.

\subsection{Primary Recovery and Error-Following Behavior}
\label{app:rq1-primary-detail}
\label{app:rq1-robustness}

Table~\ref{tab:rq1-primary-detail} reports paired comparisons on the six binary
held-out test sets, with $N=300$ examples for SMS and $N=600$ for each
remaining task. Using content-position indicators filtered under the protocol
of Appendix~\ref{app:indicator-construction}, the 16-variable J-Miner
scorecards achieve held-out source-classifier fidelity $F_J$ of
$75.0\%$--$98.3\%$ and a six-task mean of $88.8\%$, versus $F_S=71.7\%$ for the
capacity-matched Surface scorecards. The fidelity gain $\Delta F = F_J - F_S$
ranges from $+6.00$ to $+29.50$~percentage points across the six tasks, with
paired 95\% bootstrap confidence intervals above zero and exact two-sided
McNemar $p\le 1.27\times 10^{-3}$.

\begin{table}[htbp]
  \centering
  \caption{Binary recovery fidelity (\%) and paired statistical tests on the held-out test sets.}
  \label{tab:rq1-primary-detail}
  \medskip
  \begingroup
  \small
  \setlength{\tabcolsep}{2.5pt}
  \begin{tabular*}{\columnwidth}{@{\extracolsep{\fill}}lccccc@{}}
    \toprule
    Task & $F_J$ (\%) & $F_S$ (\%) & $\Delta F$ (pp) & Paired 95\% CI (pp) & McNemar $p$ \\
    \midrule
    SMS       & 98.3 & 92.3 & +6.00  & [3.00, 9.33]   & $1.21\!\times\!10^{-4}$ \\
    Sentiment & 84.3 & 62.3 & +22.00 & [17.50, 26.50] & $4.37\!\times\!10^{-20}$ \\
    Formality & 95.3 & 78.7 & +16.67 & [13.33, 20.17] & $4.52\!\times\!10^{-21}$ \\
    IMDB      & 89.5 & 69.8 & +19.67 & [15.50, 24.00] & $2.24\!\times\!10^{-18}$ \\
    Toxicity  & 90.5 & 61.0 & +29.50 & [25.33, 33.67] & $3.14\!\times\!10^{-39}$ \\
    Sarcasm   & 75.0 & 66.3 & +8.67  & [3.50, 13.67]  & $1.27\!\times\!10^{-3}$ \\
    \midrule
    Mean      & 88.8 & 71.7 & +17.08 & ---            & --- \\
    \bottomrule
  \end{tabular*}
  \endgroup
\end{table}

Table~\ref{tab:rq1-error-detail} compares the prediction-fitted scorecard $J$
with a gold-fitted control $J_{\mathrm{gold}}$ and evaluates both rules when
the source classifier disagrees with the gold label. The control
$J_{\mathrm{gold}}$ retains the 16 internal variables selected using source
predictions on $D_c$ and refits the scorecard weights to gold dataset labels.
Its full-test source-classifier fidelity $F^J_{\mathrm{gold}}$ averages $88.7\%$
equally across the six tasks, compared with $88.8\%$ for $J$.

On the subset of $n$ held-out examples where the source classifier disagrees
with the gold label, $F^J_{\mathrm{err}}$, $F^{J_{\mathrm{gold}}}_{\mathrm{err}}$,
and $F^S_{\mathrm{err}}$ measure the percentage of erroneous source decisions
reproduced by $J$, $J_{\mathrm{gold}}$, and the Surface rule $S$. Pooled across
all 281 source-error examples, $J$ reproduces 186 errors ($66.2\%$) and
$J_{\mathrm{gold}}$ reproduces 179 ($63.7\%$), whereas $S$ reproduces 143
($50.9\%$), giving an exact two-sided McNemar $p=1.5\times 10^{-4}$ between $J$
and $S$. J-Miner exceeds Surface on each of the five tasks with at least 31
source errors; on SMS, which has six source errors, J-Miner follows three and
Surface four.

\begin{table}[htbp]
  \centering
  \caption{Gold-fitted control and error-following fidelity (\%) on the held-out binary test sets.}
  \label{tab:rq1-error-detail}
  \medskip
  \small
  \setlength{\tabcolsep}{3.5pt}
  \begin{tabular*}{\columnwidth}{@{\extracolsep{\fill}}lccccc@{}}
    \toprule
    Task & $F^J_{\mathrm{gold}}$ (\%) & Error count $n$ &
    $F^{J}_{\mathrm{err}}$ (\%) & $F^{J_{\mathrm{gold}}}_{\mathrm{err}}$ (\%) & $F^{S}_{\mathrm{err}}$ (\%) \\
    \midrule
    SMS       & 98.0 & 6   & 50.0 & 50.0 & 66.7 \\
    Sentiment & 84.7 & 57  & 68.4 & 63.2 & 49.1 \\
    Formality & 95.5 & 31  & 77.4 & 77.4 & 58.1 \\
    IMDB      & 88.8 & 33  & 63.6 & 54.5 & 54.5 \\
    Toxicity  & 90.0 & 87  & 71.3 & 69.0 & 58.6 \\
    Sarcasm   & 75.2 & 67  & 55.2 & 56.7 & 35.8 \\
    \midrule
    Mean      & 88.7 & --- & --- & --- & --- \\
    Pooled    & --- & 281 & 66.2 & 63.7 & 50.9 \\
    \bottomrule
  \end{tabular*}
\end{table}

\subsection{Results Across Model Scales and Families}
\label{app:rq1-generality-detail}

We evaluate six source-model configurations across Sentiment, Formality,
IMDB, and Toxicity: Qwen3.5 at 0.8B, 2B, and 4B scales alongside Gemma-3-1B,
Llama-3.2-1B, and Phi-4-mini. Each classifier is fine-tuned and frozen under
the protocol of Appendix~\ref{app:rq1-protocol}, with J-Lens observation
windows matched at 58\%--63\% relative network depth through the penultimate
layer in Table~\ref{tab:concept-layer-windows}. Across all 24 task--classifier
cells, both J-Miner and Surface rules fix the variable budget at $K=16$, rank
candidates on construction-split predictions, and fit the same class-balanced
logistic scorecard before evaluation on the shared held-out test sets.
Table~\ref{tab:rq1-generality-detail} reports the held-out source-classifier
fidelities $F_J$ and $F_S$, downstream gold accuracies $A_J$ and $A_S$, and
95\% confidence intervals from $10{,}000$ paired bootstrap resamples for
$\Delta F = F_J - F_S$ underlying Figure~\ref{fig:result-panels}a.

\begin{table}[!ht]
  \centering
  \caption{Recovery fidelity and gold accuracy (\%) across model scales and families at $K=16$.}
  \label{tab:rq1-generality-detail}
  \medskip
  \begingroup
  \small
  \setlength{\tabcolsep}{1.2pt}
  \begin{tabular*}{\columnwidth}{@{\extracolsep{\fill}}llcccccc@{}}
    \toprule
    Task & Classifier & $F_J$ (\%) & $F_S$ (\%) & $A_J$ (\%) & $A_S$ (\%) & \shortstack{$\Delta F$\\(pp)} & \shortstack{95\% CI\\(pp)} \\
    \midrule
    Sentiment & Qwen3.5-0.8B & 84.3 & 62.3 & 80.8 & 62.5 & +22.00 & [17.50, 26.50] \\
              & Qwen3.5-2B   & 82.2 & 63.3 & 79.2 & 64.0 & +18.83 & [14.50, 23.17] \\
              & Qwen3.5-4B   & 78.0 & 64.2 & 78.8 & 63.7 & +13.83 & [9.17, 18.50] \\
              & Gemma-3-1B   & 87.5 & 63.2 & 83.0 & 64.0 & +24.33 & [20.00, 28.83] \\
              & Llama-3.2-1B & 77.3 & 62.8 & 76.2 & 63.0 & +14.50 & [9.67, 19.33] \\
              & Phi-4-mini   & 82.5 & 66.0 & 80.7 & 64.8 & +16.50 & [11.83, 21.17] \\
    \midrule
    Formality & Qwen3.5-0.8B & 95.3 & 78.7 & 92.5 & 77.8 & +16.67 & [13.33, 20.17] \\
              & Qwen3.5-2B   & 94.2 & 79.2 & 92.2 & 78.2 & +15.00 & [11.50, 18.50] \\
              & Qwen3.5-4B   & 92.2 & 80.7 & 89.2 & 78.0 & +11.50 & [8.17, 14.83] \\
              & Gemma-3-1B   & 92.2 & 80.2 & 90.3 & 78.3 & +12.00 & [8.50, 15.67] \\
              & Llama-3.2-1B & 90.7 & 78.7 & 90.3 & 78.0 & +12.00 & [8.33, 15.67] \\
              & Phi-4-mini   & 91.8 & 78.8 & 90.5 & 77.8 & +13.00 & [9.67, 16.50] \\
    \midrule
    IMDB      & Qwen3.5-0.8B & 89.5 & 69.8 & 88.0 & 69.3 & +19.67 & [15.50, 24.00] \\
              & Qwen3.5-2B   & 91.5 & 69.2 & 90.8 & 68.8 & +22.33 & [18.33, 26.33] \\
              & Qwen3.5-4B   & 88.5 & 71.7 & 89.2 & 70.7 & +16.83 & [12.83, 21.00] \\
              & Gemma-3-1B   & 88.2 & 72.3 & 87.0 & 71.8 & +15.83 & [11.83, 19.83] \\
              & Llama-3.2-1B & 90.2 & 71.2 & 89.2 & 69.8 & +19.00 & [15.17, 23.00] \\
              & Phi-4-mini   & 91.8 & 72.0 & 90.2 & 71.0 & +19.83 & [15.83, 23.83] \\
    \midrule
    Toxicity  & Qwen3.5-0.8B & 90.5 & 61.0 & 84.3 & 58.5 & +29.50 & [25.33, 33.67] \\
              & Qwen3.5-2B   & 89.3 & 61.8 & 84.2 & 58.3 & +27.50 & [22.83, 32.00] \\
              & Qwen3.5-4B   & 92.0 & 56.5 & 87.2 & 57.0 & +35.50 & [31.33, 39.83] \\
              & Gemma-3-1B   & 92.0 & 61.5 & 85.8 & 59.3 & +30.50 & [26.33, 34.67] \\
              & Llama-3.2-1B & 87.3 & 58.7 & 83.5 & 57.5 & +28.67 & [24.33, 32.83] \\
              & Phi-4-mini   & 85.0 & 58.0 & 80.8 & 55.8 & +27.00 & [22.33, 31.67] \\
    \bottomrule
  \end{tabular*}
  \endgroup
\end{table}

Across all 24 cells in Table~\ref{tab:rq1-generality-detail}, J-Miner reaches
$77.3\%$--$95.3\%$ held-out fidelity versus $56.5\%$--$80.7\%$ for the
capacity-matched Surface baseline, yielding paired gains $\Delta F$ of
$+11.50$ to $+35.50$~percentage points with all 95\% confidence intervals
above zero. Gold accuracy is also higher for J-Miner in every cell, at
$76.2\%$--$92.5\%$ versus $55.8\%$--$78.3\%$ for Surface.

\subsection{Feature Budgets and Candidate Ranking}
\label{app:parameter-sensitivity}

\begin{figure}[!ht]
  \centering
  \includegraphics[width=0.82\textwidth]{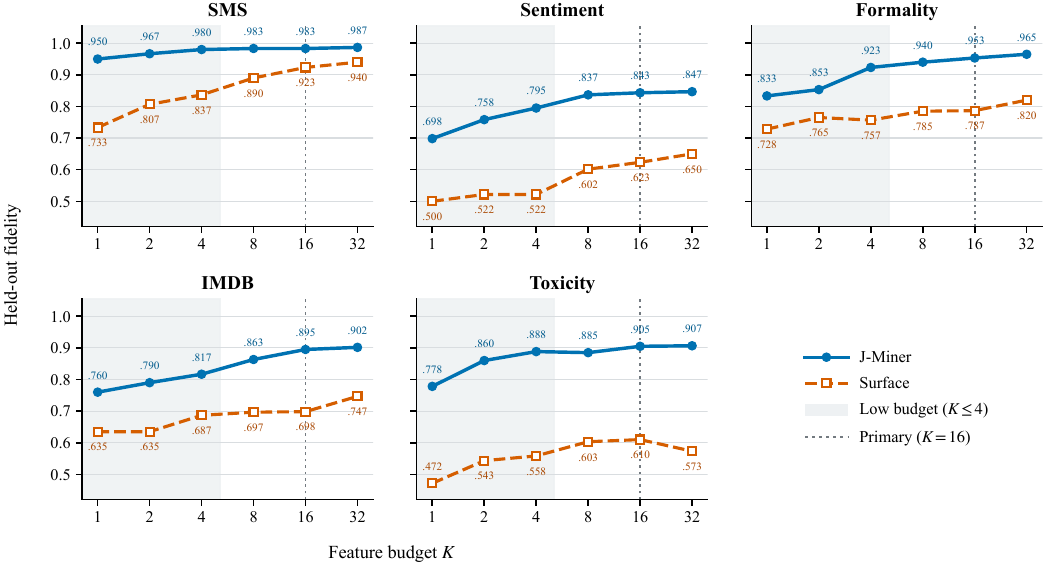}
  \caption{Source-classifier fidelity across matched feature budgets. Each panel compares
  J-Miner and Surface scorecards from $K=1$ to $K=32$ under the same source-classifier
  target and prediction head. The dotted line marks the primary $K=16$ setting.}
  \Description{Five task panels compare held-out source-classifier fidelity for J-Miner
  and Surface scorecards from one to thirty-two features.}
  \label{fig:extreme-budget}
\end{figure}

Figure~\ref{fig:extreme-budget} plots the task-level budget curves underlying
Figure~\ref{fig:compression-evidence}a from $K=1$ to $K=32$ across SMS,
Sentiment, Formality, IMDB, and Toxicity, the five binary tasks with at least
32 support-filtered surface candidates; Sarcasm yields 24 candidates under the
3\% support threshold in Appendix~\ref{app:rq1-protocol}. At $K=1$, a single
J-Miner variable with an intercept achieves $69.8\%$--$95.0\%$ fidelity, with
a five-task mean of $80.4\%$ versus $61.4\%$ for a single Surface word. Across
all 30 task-by-budget cells from $K=1$ to $K=32$, J-Miner exceeds Surface by
$+4.67$ to $+33.33$~percentage points. On SMS, Surface narrows the gap as its
vocabulary grows, rising from $73.3\%$ at $K=1$ to $94.0\%$ at $K=32$ versus
$95.0\%$ to $98.7\%$ for J-Miner. On Sentiment, IMDB, and Toxicity,
16-variable J-Miner rules reach $84.3\%$--$90.5\%$ fidelity, whereas 32-word
Surface rules reach $57.3\%$--$74.7\%$.

Table~\ref{tab:app-rq2-k-curves} compares the $\Delta(v)$-ranked prefix
against 100 random nested orderings drawn from the same support-filtered
internal candidate pool from $K=1$ to $K=64$ across all six binary tasks, with
the random-ordering median reported in parentheses. Each task's candidate pool
contains $521$ to $6{,}550$ concepts constructed without source-classifier
predictions, and every ordering is fitted and evaluated under the same LR-$K$
protocol. Across all 42 task-by-budget cells, the $\Delta(v)$-ranked prefix
exceeds the random-ordering median, with a mean gain of $20.6$~percentage
points at $K=16$ across the six tasks and $21.9$~points across the five-task
panel. On SMS, four ranked variables reach $98.0\%$ fidelity versus a
random-ordering median of $66.8\%$. Increasing the budget from $K=16$ to $K=64$
raises ranked-rule fidelity from $75.0\%$ to $82.8\%$ on Sarcasm, from
$84.3\%$ to $87.3\%$ on Sentiment, and from $89.5\%$ to $91.8\%$ on IMDB, where
the ranked prefixes remain $11.1$--$22.4$~points above the random-ordering
median at $K=64$.

\begin{table}[!ht]
  \centering
  \caption{Held-out fidelity (\%) of $\Delta(v)$-ranked prefixes and random orderings across budgets $K$.}
  \label{tab:app-rq2-k-curves}
  \medskip
  \setlength{\tabcolsep}{3.2pt}
  \begin{tabular*}{\textwidth}{@{\extracolsep{\fill}}lrrrrrrr@{}}
    \toprule
    Task & $K=1$ & $K=2$ & $K=4$ & $K=8$ & $K=16$ & $K=32$ & $K=64$ \\
    \midrule
    SMS       & 95.0 (55.3) & 96.7 (59.3) & 98.0 (66.8) & 98.3 (78.3) & 98.3 (88.3) & 98.7 (93.5) & 98.7 (96.3) \\
    Sentiment & 69.8 (55.5) & 75.8 (56.3) & 79.5 (55.9) & 83.7 (58.7) & 84.3 (63.9) & 84.7 (69.4) & 87.3 (75.0) \\
    Formality & 83.3 (52.7) & 85.3 (54.8) & 92.3 (59.9) & 94.0 (66.7) & 95.3 (76.2) & 96.5 (83.8) & 96.5 (88.3) \\
    IMDB      & 76.0 (52.2) & 79.0 (52.7) & 81.7 (53.7) & 86.3 (55.4) & 89.5 (59.0) & 90.2 (63.3) & 91.8 (69.4) \\
    Toxicity  & 77.8 (55.0) & 86.0 (54.3) & 88.8 (55.3) & 88.5 (56.4) & 90.5 (61.1) & 90.7 (68.8) & 90.8 (78.0) \\
    Sarcasm   & 58.3 (48.7) & 63.7 (52.7) & 69.5 (53.1) & 71.8 (56.8) & 75.0 (61.3) & 78.7 (66.5) & 82.8 (71.7) \\
    \bottomrule
  \end{tabular*}
\end{table}

\subsection{Observation Depths and Variable Budgets}
\label{app:dcs-grid}

\begin{figure}[!ht]
  \centering
  \includegraphics[width=0.92\textwidth]{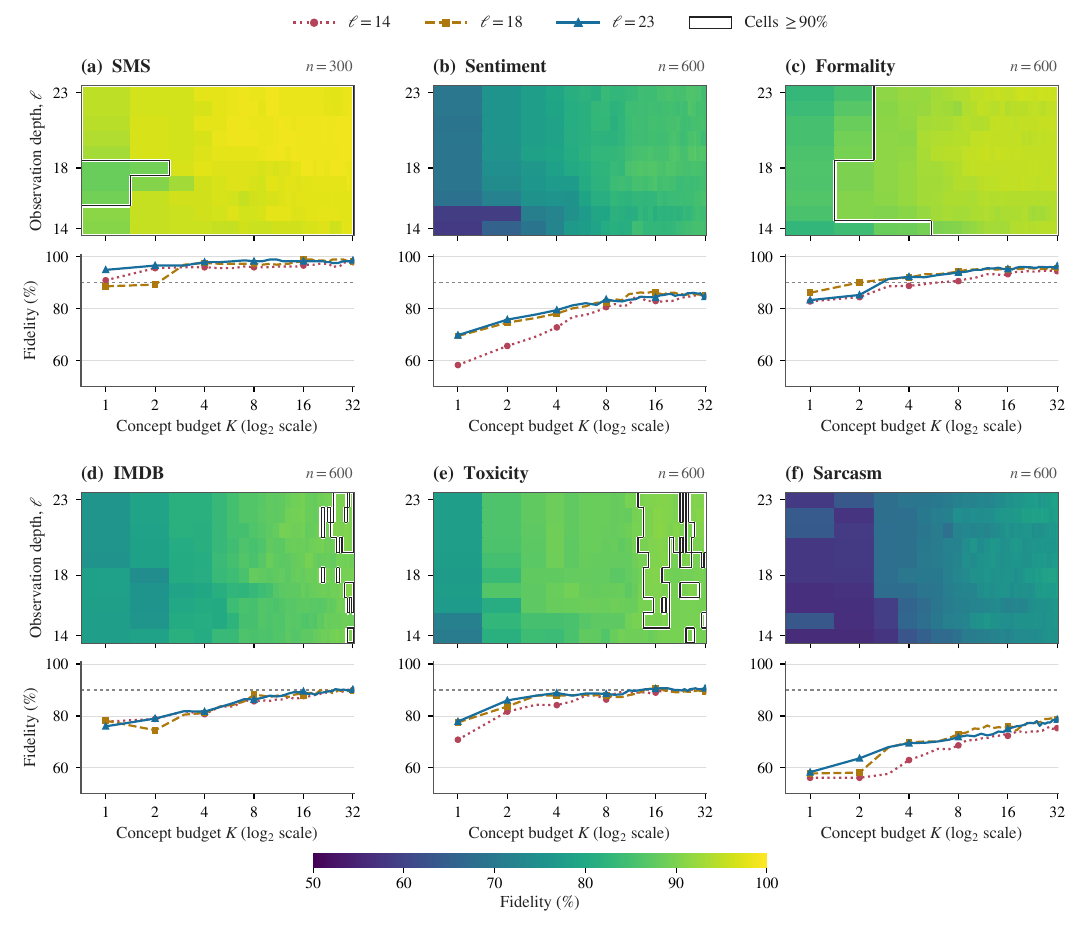}
  \caption{Held-out fidelity grids $\widehat F_{\mathcal P}(\ell,K)$
  across the six binary tasks. Upper heatmaps report held-out source-classifier
  fidelity over cumulative cutoff layers $\ell\in\{14,\dots,23\}$ and variable
  budgets $K\in\{1,\dots,32\}$, with cells at or above $90\%$ fidelity
  outlined; lower curves plot budget slices at cutoffs L14, L18, and L23.}
  \Description{Six panels, one per binary task. Each contains a heatmap of
  source-classifier fidelity over cutoff layers 14 to 23 and budgets 1 to 32 on a
  logarithmic axis, and a line plot of fidelity against budget at cutoff layers
  14, 18, and 23 with a dashed line at 90 percent.}
  \label{fig:dcs-full}
\end{figure}

Figure~\ref{fig:dcs-full} plots the held-out fidelity grids
$\widehat F_{\mathcal P}(\ell,K)$ from Equation~\eqref{eq:dcs} across cumulative
cutoff layers $\ell\in\{14,\dots,23\}$ and variable budgets $K\in\{1,\dots,32\}$
for all six binary tasks under the protocol of Appendix~\ref{app:rq1-protocol},
and Table~\ref{tab:dcs-summary} summarizes the depth comparison at $K=16$.
The full-window slice at $\ell=23$ corresponds to the primary observation window,
matching the budget sweep in Table~\ref{tab:app-rq2-k-curves} and the five-task
J-Miner budget curve in Figure~\ref{fig:compression-evidence}a.

The full-window slice in Figure~\ref{fig:dcs-full} reaches at least $90\%$
fidelity on SMS, Formality, and Toxicity at $K=16$, while IMDB reaches
$90.2\%$ at $K=32$ in Table~\ref{tab:app-rq2-k-curves}. At $K=16$,
Table~\ref{tab:dcs-summary} compares content-position readouts from layer 14
alone with the full L14--L23 window and reports the minimum and maximum
fidelity across all ten cutoffs $\ell=14,\dots,23$ on the same held-out test
set per task. Rules using L14 alone reach $83.0\%$, $87.7\%$, and $89.0\%$
fidelity on Sentiment, IMDB, and Toxicity, within $1.3$--$1.8$~percentage
points of the full-window results, and $K=16$ fidelity varies by
$1.8$--$4.8$~points across all ten cutoffs.

\begin{table}[!ht]
  \centering
  \caption{Held-out source-classifier fidelity (\%) at $K=16$ across observation depth cutoffs.}
  \label{tab:dcs-summary}
  \medskip
  \begingroup
  \small
  \setlength{\tabcolsep}{3pt}
  \begin{tabular*}{\linewidth}{@{\extracolsep{\fill}}lrccc@{}}
    \toprule
    Task & Test $n$ & L14 (\%) & L23 (\%) & Range across cutoffs (\%) \\
    \midrule
    SMS       & 300 & 96.7 & 98.3 & 96.7--99.0 \\
    Sentiment & 600 & 83.0 & 84.3 & 81.5--86.3 \\
    Formality & 600 & 93.3 & 95.3 & 93.3--95.7 \\
    IMDB      & 600 & 87.7 & 89.5 & 86.5--89.5 \\
    Toxicity  & 600 & 89.0 & 90.5 & 89.0--90.8 \\
    Sarcasm   & 600 & 72.3 & 75.0 & 72.3--76.7 \\
    \bottomrule
  \end{tabular*}
  \endgroup
\end{table}

The held-out label-readout trajectories in Figure~\ref{fig:compression-evidence}b
provide a reference for these depth results. At each layer, verdict agreement
is the fraction of test inputs whose terminal \texttt{Decision:} label-margin
sign matches the source classifier's final prediction. L95 is the earliest
layer from which this agreement remains at least $95\%$ through L22. L95
occurs at L10 for SMS, L14 for Formality, and L15 for Sentiment, IMDB,
Toxicity, and Sarcasm. At L14, direct label agreement on Sentiment, IMDB, and
Toxicity is $60.3\%$, $54.7\%$, and $55.8\%$, respectively. Together with the
L14 scorecard results in Table~\ref{tab:dcs-summary}, these trajectories show
that content-position readouts support compact rule recovery at a depth where
direct label readout has not yet reached sustained $95\%$ agreement.
\FloatBarrier

\section{Inspecting the Recovered Interface}
\label{app:interface-inspection}
\label{app:rq2-reveal}

This appendix extends the analysis of internal evidence and rule composition in
\hyperref[sec:rq2-reveal]{RQ2}. We examine literal versus non-literal variable
activations, sensitivity to structural readout positions, linear scorecard
composition, targeted word-deletion interventions, and alternative internal
observation protocols within the EDC framework.

\subsection{Non-Literal Activation and Positional Robustness}
\label{app:rq2-nonliteral}

In Equation~\eqref{eq:app-message-indicator}, the message-level indicator
$c_v(x)\in\{0,1\}$ records whether vocabulary coordinate $v\in\mathcal V$
enters the top-$R$ J-Lens readouts across intermediate hidden states. Using
case-normalized substring matching for the printed name of $v$ in input $x$, we
decompose $c_v(x)$ and its construction-split signed trigger-rate gap
$\Delta_{\mathrm{sgn}}(v)=\pi_1(v)-\pi_0(v)$ into literal and non-literal parts:
\begin{equation}
  c_v(x) = c_v^{\mathrm{lit}}(x) + c_v^{\mathrm{non}}(x),
  \qquad
  \Delta_{\mathrm{sgn}}(v) = \Delta^{\mathrm{lit}}(v) + \Delta^{\mathrm{non}}(v),
  \label{eq:nonliteral-decomp}
\end{equation}
where $c_v^{\mathrm{lit}}(x)=1$ if and only if $c_v(x)=1$ and the printed name
of $v$ occurs in $x$. Table~\ref{tab:app-rq2-nonliteral-decision} shows that
literal string matches account for 1.5\%--27.0\% of variable activations across
the six binary tasks, with a six-task mean of 9.8\%. Under the decomposition in
Equation~\eqref{eq:nonliteral-decomp}, the non-literal component supplies
72.6\%--99.1\% of the total class-discriminative gap
$\sum_{v\in\mathcal V}\operatorname{sgn}(\Delta_{\mathrm{sgn}}(v))\,\Delta^{\mathrm{non}}(v)\allowbreak/\allowbreak\sum_{v\in\mathcal V}|\Delta_{\mathrm{sgn}}(v)|$,
averaging 90.5\% across tasks. Holding the 16 selected variables fixed and
refitting the linear scorecards on $D_c$ with non-literal activations
$c_v^{\mathrm{non}}$ alone yields 88.6\% mean held-out fidelity, compared with
88.8\% for the full-activation scorecard and 71.7\% for the 16-word Surface
baseline, exceeding Surface by 16.9~pp. Restricting the same variables to
literal activations $c_v^{\mathrm{lit}}$ reduces mean fidelity to 63.3\%,
dropping to 49.3\% on Sentiment and 45.8\% on Toxicity.

\begin{table}[htbp]
  \centering
  \caption{Literal and non-literal activation shares (\%) and refitted LR-16
  held-out fidelity (\%) across the six binary tasks.}
  \label{tab:app-rq2-nonliteral-decision}
  \medskip
  \small
  \begin{tabular*}{\linewidth}{@{\extracolsep{\fill}}lrrrrrr@{}}
    \toprule
    & \multicolumn{2}{c}{Share (\%)} & \multicolumn{4}{c}{Held-out fidelity (\%)} \\
    \cmidrule(lr){2-3}\cmidrule(lr){4-7}
    Task & Non-literal gap & Literal activations & All & Literal & Non-literal & Surface \\
    \midrule
    SMS       & 94.8 & 5.8  & 98.3 & 76.7 & 98.7 & 92.3 \\
    Sentiment & 95.6 & 4.1  & 84.3 & 49.3 & 84.3 & 62.3 \\
    Formality & 72.6 & 27.0 & 95.3 & 83.5 & 94.0 & 78.7 \\
    IMDB      & 91.3 & 6.6  & 89.5 & 65.7 & 88.5 & 69.8 \\
    Toxicity  & 99.1 & 1.5  & 90.5 & 45.8 & 90.7 & 61.0 \\
    Sarcasm   & 89.8 & 13.5 & 75.0 & 58.8 & 75.3 & 66.3 \\
    \midrule
    Mean      & 90.5 & 9.8  & 88.8 & 63.3 & 88.6 & 71.7 \\
    \bottomrule
  \end{tabular*}
\end{table}

To test whether rule fidelity depends on boundary or punctuation positions, we
recompute each test input's message-level state $c_\theta(x)$ after excluding
selected position classes from the L14--L23 event union. The input and its
hidden states remain unchanged, and the frozen LR-16 scorecard is evaluated
without refitting. In Table~\ref{tab:structural-positions}, excluding positions
that contain only sentence-ending punctuation (\texttt{.}, \texttt{!},
\texttt{?}) lowers held-out fidelity from 84.3\% to 82.5\% in Sentiment and
from 90.5\% to 85.3\% in Toxicity. Excluding the final content position gives
82.3\% and 85.5\%, respectively. When all token positions without letters or
digits are excluded from pooling, fidelity remains at 80.5\% and 84.7\%, well
above the full-input Surface baselines of 62.3\% and 61.0\%.

\begin{table}[htbp]
  \centering
  \caption{Held-out fidelity (\%) of frozen LR-16 scorecards after excluding
  structural or boundary readout positions prior to pooling.}
  \label{tab:structural-positions}
  \medskip
  \small
  \begin{tabular*}{\linewidth}{@{\extracolsep{\fill}}lcc@{}}
    \toprule
    Excluded positions & Sentiment (\%) & Toxicity (\%) \\
    \midrule
    None, full content sequence & 84.3 & 90.5 \\
    Sentence-ending punctuation only & 82.5 & 85.3 \\
    Final content position & 82.3 & 85.5 \\
    All tokens without letters or digits & 80.5 & 84.7 \\
    \midrule
    Surface rule, full input & 62.3 & 61.0 \\
    \bottomrule
  \end{tabular*}
\end{table}

\subsection{Composition of Recovered Rules}
\label{app:rq2-binary-trace}

\begin{figure*}[htbp]
  \centering
  \includegraphics[width=\textwidth]{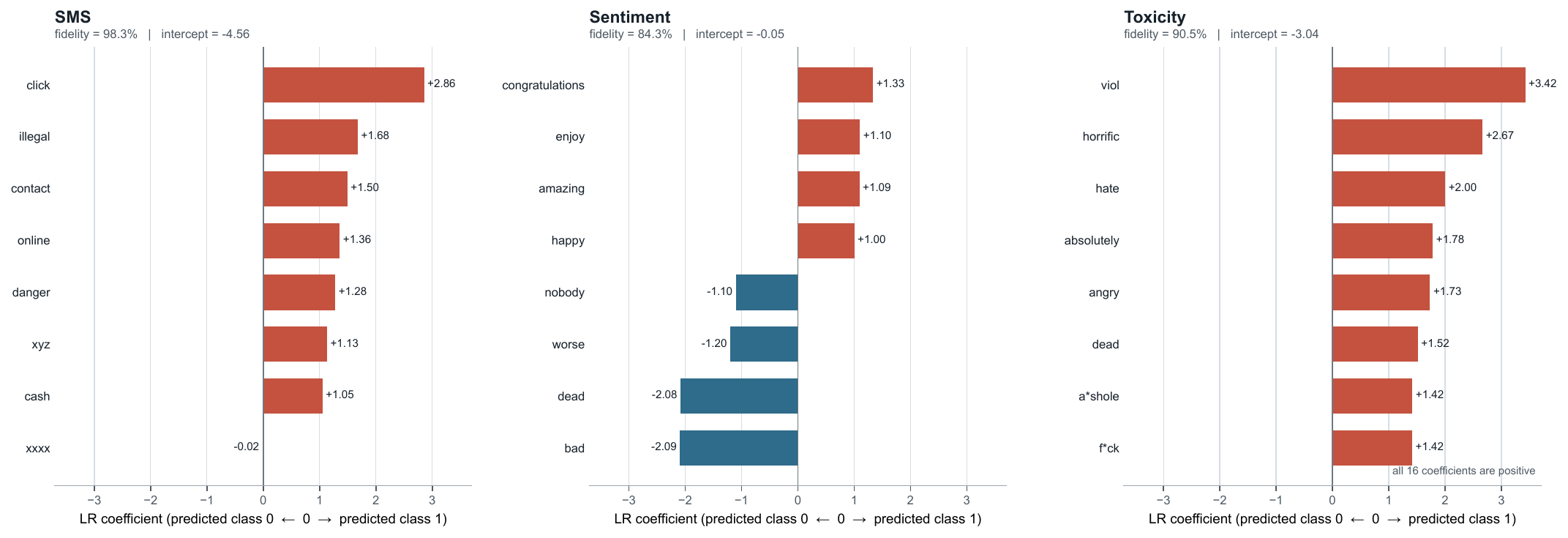}
  \caption{Recovered 16-variable linear scorecards for SMS, Sentiment, and
  Toxicity on Qwen3.5-0.8B, showing eight representative coefficients per task.}
  \Description{Three horizontal bar charts show signed weights for eight
  selected coordinates in the SMS, Sentiment, and Toxicity scorecards.}
  \label{fig:rq1-scorecards}
\end{figure*}

Under Equation~\eqref{eq:edc-scorecard}, the frozen linear scorecard evaluates
$z(c_\theta(x))=b+\sum_{j=1}^K w_j c_{\theta,j}(x)$ on the extracted state
$c_\theta(x)\in\{0,1\}^K$. The intercept $b$ sets the default score when no
variable is active, and each active coordinate contributes a signed evidence
term $a_j(x)=w_j c_{\theta,j}(x)$. Figure~\ref{fig:rq1-scorecards} shows how
these signed contributions combine in the 16-variable scorecards on
Qwen3.5-0.8B. In SMS and Toxicity, negative intercepts establish non-spam and
non-toxic defaults, while activating positively weighted variables such as
\texttt{click} and \texttt{cash} in SMS or \texttt{absolutely} and \texttt{hate}
in Toxicity shifts the score toward the positive class. The Sentiment scorecard
combines positive weights on \texttt{amazing} and \texttt{happy} with negative
weights on \texttt{bad}, \texttt{dead}, and \texttt{worse}, allowing co-active
variables to reinforce or offset one another in the final score.

Across the six binary tasks, 95 of the 96 selected variables, or 99.0\%, have
jointly fitted weights whose signs match their construction-split marginal gaps
$\Delta_{\mathrm{sgn}}(v)=\pi_1(v)-\pi_0(v)$. All 96 variables also retain
their construction-split marginal gap signs on the held-out test sets, with
each variable's pointwise 95\% bootstrap confidence interval over $2{,}000$
resamples excluding zero in that direction.

\subsection{Rule-Guided Text Edits}
\label{app:rq2-text-edits}

To evaluate whether input words localized by J-Lens readouts are behaviorally
relevant to the source classifier, we perform paired word deletions using the
frozen Qwen3.5-0.8B classifiers and LR-16 scorecards. On each input with active
rule evidence, we select the active variable $j^\star$ with the largest absolute
contribution $|w_j c_{\theta,j}(x)|$, breaking ties by variable index. Writing
the source logit margin as $m(x)=\mathrm{logit}_1(x)-\mathrm{logit}_0(x)$, we
define the pre-edit predicted direction and its agreement with the response to
an edited input $x'$ as
\begin{equation}
  \begin{aligned}
  d_{\mathrm{pred}}(x)&=\operatorname{sign}(-w_{j^\star}c_{\theta,j^\star}(x)),\\
  A(x,x')&=\mathbf 1\big[d_{\mathrm{pred}}(x)(m(x')-m(x))>10^{-6}\big].
  \end{aligned}
\end{equation}
Margin changes within $10^{-6}$ count as non-agreement, and
$d_{\mathrm{pred}}(x)$ is fixed before editing.

Each targeted edit collects all content positions where variable $j^\star$
enters the top $R=3$ vocabulary coordinates across layers L14--L23, maps those
positions via tokenizer offsets to complete input words, and removes their
deduplicated union. We retain eligible inputs ($n/N$ in
Table~\ref{tab:rq2-text-deletion}) that have at least one active variable, a
valid complete-word mapping, untruncated text within the model's context limit,
and a deletion span covering less than half the input words. The prompt
template and remaining words stay unchanged.

\begin{figure}[htbp]
  \centering
  \includegraphics[width=\linewidth]{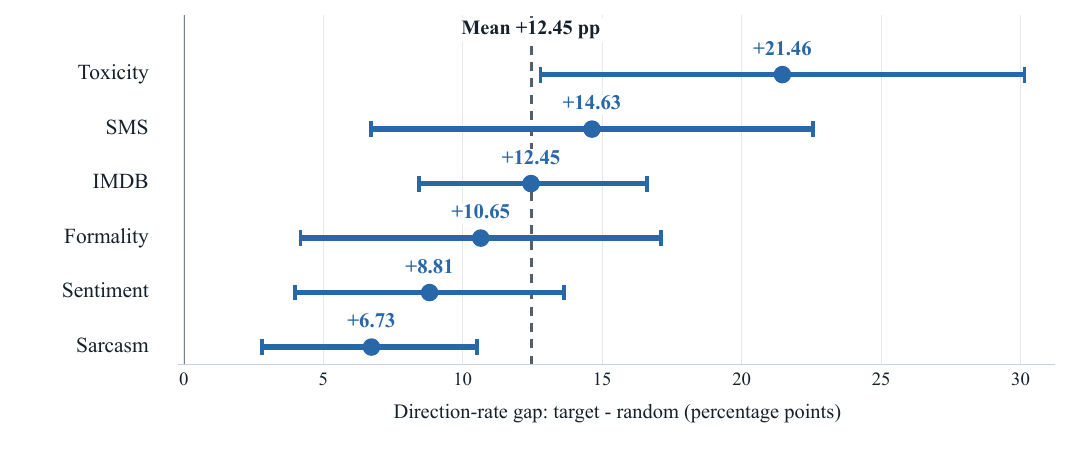}
  \caption{Direction-agreement gap between rule-guided and matched-random word
  deletions across the six binary classification tasks. Points report
  percentage-point gains in the rate at which edits shift the source
  classifier's margin in the rule-predicted direction, with horizontal bars
  showing paired 95\% confidence intervals and the vertical dashed line
  marking the six-task mean of $+12.45$~pp. Absolute rates and eligible sample
  counts appear in Table~\ref{tab:rq2-text-deletion}.}
  \label{fig:rq2-text-deletion-gap}
  \Description{Direction-agreement gaps are 21.46 points for Toxicity, 14.63
  for SMS, 12.45 for IMDB, 10.65 for Formality, 8.81 for Sentiment, and 6.73
  for Sarcasm. A dashed vertical line marks the equal-task mean of 12.45 points.}
\end{figure}

Figure~\ref{fig:rq2-text-deletion-gap} and Table~\ref{tab:rq2-text-deletion}
compare these targeted deletions against random controls that remove the same
number of words from the same input, including possible overlap with the
targeted words. Agreement is averaged within each input before task-level
aggregation, and paired 95\% confidence intervals are computed over $5{,}000$
bootstrap resamples.

Across the six binary tasks, direction-agreement gains over random deletion
range from $+6.73$~pp to $+21.46$~pp and average $+12.45$~pp, with every paired
95\% confidence interval strictly above zero.

\begin{table}[htbp]
  \centering
  \caption{Direction-agreement rates (\%) and paired differences (pp) for
  rule-guided versus matched-random word deletions.}
  \label{tab:rq2-text-deletion}
  \medskip
  \small
  \setlength{\tabcolsep}{3pt}
  \begin{tabular*}{\linewidth}{@{\extracolsep{\fill}}lrrrr@{}}
    \toprule
    & & \multicolumn{2}{c}{Agreement (\%)} & \\
    \cmidrule(lr){3-4}
    Task & Eligible $n/N$ & Targeted & Random & Gap [95\% CI] \\
    \midrule
    SMS       & 164/300 & 73.78 & 59.15 & +14.63 [6.71, 22.56] \\
    Sentiment & 401/600 & 55.61 & 46.80 & +8.81 [3.99, 13.63] \\
    Formality & 263/600 & 57.03 & 46.39 & +10.65 [4.18, 17.11] \\
    IMDB      & 482/600 & 51.45 & 39.00 & +12.45 [8.44, 16.60] \\
    Toxicity  & 219/600 & 59.36 & 37.90 & +21.46 [12.79, 30.14] \\
    Sarcasm   & 545/600 & 68.44 & 61.71 & +6.73 [2.81, 10.52] \\
    \bottomrule
  \end{tabular*}
\end{table}
\FloatBarrier

\subsection{EDC with Alternative Observation Protocols}
\label{app:rq2-interfaces}

In Sections~\ref{sec:edc-objective} and~\ref{sec:edc-mining}, the observation
protocol $\mathcal O$ specifies how intermediate hidden states are mapped into
candidate binary events before variable selection and rule fitting. We compare
16-variable EDC rules on Toxicity, Formality, Sentiment, and IMDB across three
internal observation protocols: J-Lens, vanilla logit-lens readout, and
task-trained sparse autoencoder (SAE) features. Both J-Lens and vanilla readout
project intermediate residual states from the layer windows in
Table~\ref{tab:concept-layer-windows} into the source classifier's vocabulary
space and retain the top $R=3$ coordinates at each content position. J-Lens
applies the external WikiText-103 transport operators $J_\ell$ before final
normalization and unembedding as in Equation~\eqref{eq:vocabulary-readout},
whereas vanilla readout applies $U\,\mathrm{Norm}(h_{\ell,p}(x))$ directly to
each untransported residual state. For the SAE protocol, we train a separate
Top-$K$ sparse autoencoder with $16{,}384$ latents and active sparsity $k=32$
on each task's unit-normalized construction-split residual states across content
positions of layers 14--22. Each autoencoder is trained for 10 epochs with Adam
at learning rate $10^{-3}$ and batch size 256, and a latent fires whenever its
activation at any content position reaches at least $50\%$ of its
construction-split maximum. All three protocols share the $3\%$ minimum support
filter, trigger-rate gap ranking $\Delta(v)$, the $K=16$ variable budget, and
the class-balanced logistic scorecard fitting objective on $D_c$.

Table~\ref{tab:interfaces} reports held-out fidelity to the frozen source
classifier under each protocol. Four-task mean fidelities are 89.92\% for
J-Lens, 90.83\% for vanilla readout, and 91.71\% for task-trained SAE features,
compared with 67.96\% for the 16-word Surface rule. All three internal
protocols outperform Surface on every task, showing that small-budget recovery
holds across different internal observation interfaces. Unlike task-trained SAEs,
J-Lens and vanilla readout require no per-task dictionary training and index
variables directly in the classifier's vocabulary rather than unlabeled latent
coordinates. Between the two vocabulary protocols, calibrating intermediate
states with $J_\ell$ aligns the selected coordinates with the task language.
Defining complete English words as lowercase $a$--$z$ strings present in the
Unix \texttt{web2} dictionary or occurring as complete tokens in the six binary
datasets, 63 of the 64 selected J-Lens coordinates are complete English words
with zero subword fragments. By contrast, uncalibrated vanilla readout selects
24 complete English words alongside 29 non-English script tokens and 11 subword
fragments such as \texttt{erci}, \texttt{pekt}, and \texttt{lessly}. We
therefore use J-Lens as the default readout when matching internal activations
against English input words in
Appendices~\ref{app:rq2-nonliteral}--\ref{app:rq2-text-edits}.

\begin{table}[htbp]
  \centering
  \caption{Held-out source-classifier fidelity (\%) and complete English word
  counts for 16-variable linear scorecards across internal observation
  protocols.}
  \label{tab:interfaces}
  \medskip
  \small
  \begin{tabular*}{\linewidth}{@{\extracolsep{\fill}}lrrrrc@{}}
    \toprule
    & \multicolumn{3}{c}{Observation protocol within EDC (\%)} & Input reference (\%) & English words \\
    \cmidrule(lr){2-4}\cmidrule(lr){5-5}\cmidrule(l){6-6}
    Task & J-Lens & Vanilla & SAE & Surface & J-Lens / Vanilla \\
    \midrule
    Toxicity  & 90.50 & 91.83 & 93.33 & 61.00 & 16 / 3 \\
    Formality & 95.33 & 96.33 & 97.83 & 78.67 & 15 / 10 \\
    Sentiment & 84.33 & 86.17 & 81.67 & 62.33 & 16 / 6 \\
    IMDB      & 89.50 & 89.00 & 94.00 & 69.83 & 16 / 5 \\
    \midrule
    Mean / Total & 89.92 & 90.83 & 91.71 & 67.96 & 63 / 24 \\
    \bottomrule
  \end{tabular*}
\end{table}
\FloatBarrier

\section{Standalone Rule Execution and Reader Replacement}
\label{app:reader-replacement}
\label{app:reader-replacement-protocol}
\label{app:reader-replacement-decomposition}
\label{app:reader-task-performance}

Standalone execution in \hyperref[sec:rq3-transfer]{RQ3} replaces the source
concept states with text-reader predictions while keeping the 16 variables and
the recovered LR-16 scorecard $g$ fixed. J-only, J+Aux, and Plain KD use the
checkpoints selected under the protocol in
Appendix~\ref{app:student-reconstruction}, with all predictions and statistical
tests in Tables~\ref{tab:transfer}--\ref{tab:reader-replacement} and
Figure~\ref{fig:correction-harm}c paired across the 3{,}300 held-out test
inputs.

All source states $c=c_\theta(x)$ and reader states
$\widetilde c=\widetilde c_\phi(x)$ yield nonzero scorecard logits on these
inputs, so the decision-change decomposition in
Proposition~\ref{prop:edc-direction} applies without ties. By
Proposition~\ref{prop:edc-preservation-formal}, the margin condition
$E_w^-(c,\widetilde c)<\gamma(c)$ certifies a subset of preserved decisions,
giving $\mathrm{Cover}\le\mathrm{Keep}$. The changed decisions partition into
corrections and harms relative to the source classifier, with
$1-\mathrm{Keep}=p_{\mathrm{corr}}+p_{\mathrm{harm}}$ and
$F_\phi-F_\Pi=p_{\mathrm{corr}}-p_{\mathrm{harm}}$.

Table~\ref{tab:reader-replacement} relates these decision changes to the
direction of the score shift. Toward is the fraction of inputs with a nonzero
score change on which $\Delta_Y=(2Y-1)(z(\widetilde c)-z(c))>0$, where
$Y=f_\theta(x)$. Across the six tasks, Toward averages $86.1\%$ under J-only
and $91.0\%$ under J+Aux, with 95\% bootstrap CIs of $[84.7\%, 87.5\%]$ and
$[89.8\%, 92.1\%]$. Mean harm rates are $4.78\%$ and $3.86\%$, yielding net
fidelity gains of $+2.22$~pp and $+4.42$~pp, respectively.

\begin{table}[htbp]
  \centering
  \caption{Decision-change decomposition and directional alignment under standalone reader replacement.}
  \label{tab:reader-replacement}
  \medskip
  \small
  \setlength{\tabcolsep}{6pt}
  \begin{tabular*}{\linewidth}{@{\extracolsep{\fill}}llrrrr@{}}
    \toprule
    Task & Reader & $p_{\mathrm{corr}}$ (\%) & $p_{\mathrm{harm}}$ (\%)
         & $F_\phi-F_\Pi$ (pp) & Toward (\%) \\
    \midrule
    SMS       & J-only & 0.33 & 1.33 & $-1.00$ & 91.3 \\
              & J+Aux  & 1.00 & 0.67 & $+0.33$ & 94.9 \\
    Sentiment & J-only & 11.50 & 5.67 & $+5.83$ & 89.4 \\
              & J+Aux  & 12.17 & 4.50 & $+7.67$ & 90.9 \\
    Formality & J-only & 2.17 & 5.17 & $-3.00$ & 88.1 \\
              & J+Aux  & 3.17 & 3.00 & $+0.17$ & 94.9 \\
    IMDB      & J-only & 6.33 & 3.17 & $+3.17$ & 89.4 \\
              & J+Aux  & 7.67 & 3.00 & $+4.67$ & 92.6 \\
    Toxicity  & J-only & 4.17 & 5.67 & $-1.50$ & 81.6 \\
              & J+Aux  & 4.83 & 5.50 & $-0.67$ & 84.0 \\
    Sarcasm   & J-only & 17.50 & 7.67 & $+9.83$ & 77.0 \\
              & J+Aux  & 20.83 & 6.50 & $+14.33$ & 88.6 \\
    \midrule
    Mean      & J-only & 7.00 & 4.78 & $+2.22$ & 86.1 \\
              & J+Aux  & 8.28 & 3.86 & $+4.42$ & 91.0 \\
    \bottomrule
  \end{tabular*}
\end{table}

The correction rate is bounded by the initial rule disagreement,
$p_{\mathrm{corr}}\le 1-F_\Pi$. On SMS, Formality, and Toxicity, source-state
fidelity is $90.5\%$--$98.3\%$, leaving $1.7\%$--$9.5\%$ of test decisions
available for correction. J+Aux stays within $0.67$~pp of source-state fidelity
on these tasks, with exact two-sided McNemar $p\ge .70$. On Sentiment, IMDB, and
Sarcasm, source-state fidelity is $84.3\%$, $89.5\%$, and $75.0\%$, and
J+Aux improves it by $+7.67$~pp, $+4.67$~pp, and $+14.33$~pp, respectively,
with $p<.001$ on each task. In Table~\ref{tab:transfer}, paired comparisons with
Plain KD detect no significant fidelity difference on any task ($p\ge .146$),
with six-task means of $93.3\%$ for J+Aux and $92.8\%$ for Plain KD.

\end{document}